\documentclass[10pt,twocolumn,letterpaper]{article}
\usepackage[hmargin=0.8125in,top=1in,textheight=8.875in]{geometry}
\usepackage{times,xspace,graphicx,amsmath,amssymb,booktabs}
\usepackage[dvipsnames]{xcolor}
\usepackage[numbers,sort&compress]{natbib}
\usepackage[font=small,labelsep=period,skip=6pt]{caption}
\usepackage[shortlabels,inline]{enumitem}
\usepackage{multirow,makecell,cuted,colortbl,placeins,longtable,array}
\usepackage{algorithm}
\usepackage[noend]{algpseudocode}
\usepackage{microtype}
\usepackage{titling}
\usepackage[skins]{tcolorbox}

\setlist[itemize,enumerate]{noitemsep,leftmargin=*,topsep=0pt}

\definecolor{cvprblue}{rgb}{0.21,0.49,0.74}
\definecolor{abstractfill}{HTML}{DEF5FD}
\usepackage[pagebackref,breaklinks,colorlinks,allcolors=cvprblue]{hyperref}
\usepackage[capitalize]{cleveref}

\newcommand{\method}{\textsc{ViTrust}\xspace}

\newcommand{\vect}[1]{\mathbf{#1}}
\newcommand{\best}[1]{\textbf{#1}}

\definecolor{bestrow}{HTML}{E6F2FA}
\definecolor{tableoneblue}{HTML}{E6F2FA}
\definecolor{reconstructiongreen}{HTML}{E8F3E8}
\definecolor{predictionpink}{HTML}{FBE8EE}
\Crefname{section}{Sec.}{Secs.}
\crefname{section}{Sec.}{Secs.}
\Crefname{table}{Tab.}{Tabs.}
\crefname{table}{Tab.}{Tabs.}
\Crefname{figure}{Fig.}{Figs.}
\crefname{figure}{Fig.}{Figs.}

\RenewDocumentEnvironment{abstract}{+b}{%
  \begin{tcolorbox}[
    enhanced,
    colback=abstractfill,
    colframe=abstractfill,
    boxrule=0pt,
    arc=9pt,
    drop fuzzy shadow south=black!26,
    boxsep=0pt,
    left=16pt, right=16pt, top=10pt, bottom=11pt,
    before skip=0pt, after skip=0pt,
  ]
    {\sffamily\bfseries\large\abstractname\par}
    \smallskip
    \small\noindent#1\par
  \end{tcolorbox}
}{}

\pretitle{\centering\fontsize{20}{23}\selectfont\bfseries}
\posttitle{\par\vskip 12pt}
\preauthor{\centering\normalsize\begin{tabular}{@{}c@{}}}
\postauthor{\end{tabular}\par}
\predate{}
\postdate{}

\title{ViTrust: Learning 3D Observation Trust\\
for Deformable Physical World Models}
\author{%
  Zhihao Guo\textsuperscript{1}\enspace
  Zhihao Li\textsuperscript{1}\enspace
  Xu Ge\textsuperscript{1}\enspace
  Lijian Zhong\textsuperscript{1}\enspace
  Yu Lu\textsuperscript{1,2}\enspace
  Haitao Wu\textsuperscript{1,2}\enspace
  Longqing Zhao\textsuperscript{1,3}\enspace
  Yue Ma\textsuperscript{1,}\thanks{Corresponding author.}\\[5pt]
  {\small\color{black!78}%
    \textsuperscript{1}\,Hesai Technology\qquad
    \textsuperscript{2}\,Shanghai Jiao Tong University\qquad
    \textsuperscript{3}\,Southeast University}
}
\date{}
\hypersetup{
  pdftitle={ViTrust: Learning 3D Observation Trust for Deformable Physical World Models},
  pdfauthor={Zhihao Guo, Zhihao Li, Xu Ge, Lijian Zhong, Yu Lu, Haitao Wu, Longqing Zhao, Yue Ma}
}

\begin{document}
\maketitle

\begin{strip}
  \vspace{-19pt}
  \begin{abstract}
Learning deformable physical world models requires dynamics that generalize
beyond observed motion. Accurate reconstruction alone does not ensure
accurate future prediction, motivating us to learn how observations
should influence physical identification.
However, visual reliability alone does not determine how informative
an observation is about physical dynamics.
We propose \method to learn 3D observation trust, determining how strongly
each tracked point influences physical identification.
Its key design is a visual--contact trust adapter that combines
pointwise spatiotemporal evidence with interaction history on a persistent
contact graph to predict time-varying observation weights.
A single-stage, shared-forward objective jointly optimizes these weights
and physical parameters through weighted geometry and correspondence
losses, with constraints to prevent observation-weight collapse.
All learning uses only the observed prefix, fitted physics remains fixed
for future prediction under supplied controls.
Experiments on real-world sequences demonstrate improved future prediction over the evaluated baselines, alongside qualitative demonstrations of novel-control rollouts and cross-object transfer.
\end{abstract}

  \vspace{8pt}
  \centering
  \includegraphics[width=\textwidth]{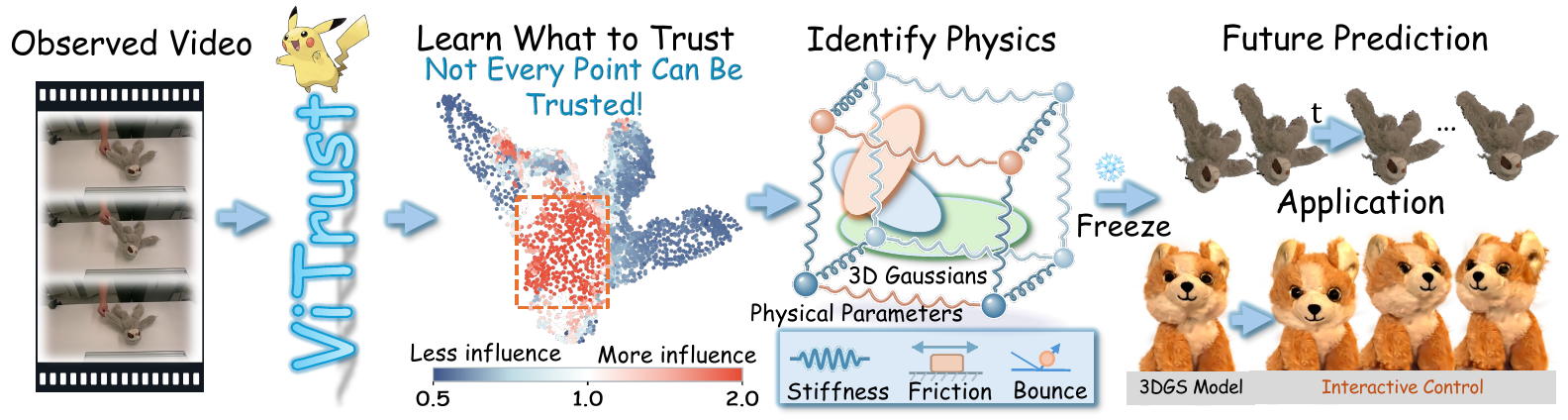}
  \captionsetup{hypcap=false,labelfont=bf}
  \captionof{figure}{\textbf{Learning 3D observation trust for deformable physical world models.}
  \method learns how strongly each tracked 3D point should influence physical
  identification, beyond visual confidence alone. Spatiotemporal visual evidence
  and contact history on a persistent graph inform time-varying observation
  weights (blue: low; red: high). A \textbf{single-stage joint optimization}
  learns these weights together with physical parameters through weighted
  geometry and tracking errors, with constraints to prevent observation-weight
  collapse. The fitted physics is then frozen for future prediction and
  transferred to a reconstructed plush dog~\cite{smartdatascan2025plushdog}
  under novel controls, without target-motion fitting.}
  \label{fig:teaser}
\end{strip}

\raggedbottom

\section{Introduction}

World models predict how an environment responds
to actions, beyond reproducing observations
~\cite{ha2018worldmodels,zhou2025dinowm,yang2025physworld}.
We study deformable physical world models that identify an explicit simulator
from a short RGB-D interaction. From a cloth lift or plush manipulation,
these models infer material and contact parameters to predict responses to
subsequent controls. This leads to our central question:
\textcolor{cvprblue}{which observations should guide physical identification
for future prediction?}

Existing approaches learn visual or particle dynamics, as in DINO-WM and
GS-Dynamics~\cite{zhou2025dinowm,zhang2024gsdynamics}, or fit explicit
simulators to video, as in PAC-NeRF, Spring-Gaus, and
PhysTwin~\cite{li2023pacnerf,zhong2024springgaus,jiang2025phystwin}.
MatPhys~\cite{yang2026matphys} introduces reusable material priors, while
DeformMaster and PhysCoRe~\cite{li2026deformmaster,yin2026physcore} combine
physics with learned dynamics corrections. These advances improve the
models used to predict motion. A complementary issue is how observations
should influence physical identification: richer dynamics do not by
themselves determine which measurements should dominate fitting. For example,
PhysTwin masks invalid tracks but assigns equal fitting weights to valid points,
without distinguishing their measurement quality or identification value.

Reliable measurements need not be physically
informative: a well-tracked region undergoing little deformation may provide
limited evidence about stiffness. Motion near contact can reveal deformation
under an applied control, but tracking errors there can also bias the fit.
Equal weighting can let numerous weakly informative tracks dominate, while
measurement confidence alone does not capture their identification value.
Moreover, \textcolor{cvprblue}{\textbf{a small reconstruction residual alone
does not establish that an observation is informative about the dynamics}}:
it can reflect easy-to-fit
motion, while a large residual can arise from either measurement errors or
simulator mismatch. Learning observation influence therefore requires both
visual--contact evidence and constraints that prevent the adapter from
trivially suppressing difficult measurements.

We introduce \method to learn 3D observation trust, the relative influence of
each tracked point at each time on physical identification. Local
spatiotemporal visual evidence describes observation consistency, while a
persistent contact graph relates motion to the evolving interaction. An
adapter combines these cues into time-varying observation weights.
Starting from a shared physical initialization, we jointly optimize the
adapter and physical parameters with weighted surface and correspondence
errors, regularized against weight concentration and temporal instability.
During joint training, each simulated frame provides both
partial gradients before either parameter set is updated, without candidate
generation. All updates use the observed
prefix, and the final physical parameters remain fixed during future
prediction without a separate refitting stage.

\begin{figure}[t]
  \centering
  \includegraphics[width=\columnwidth]{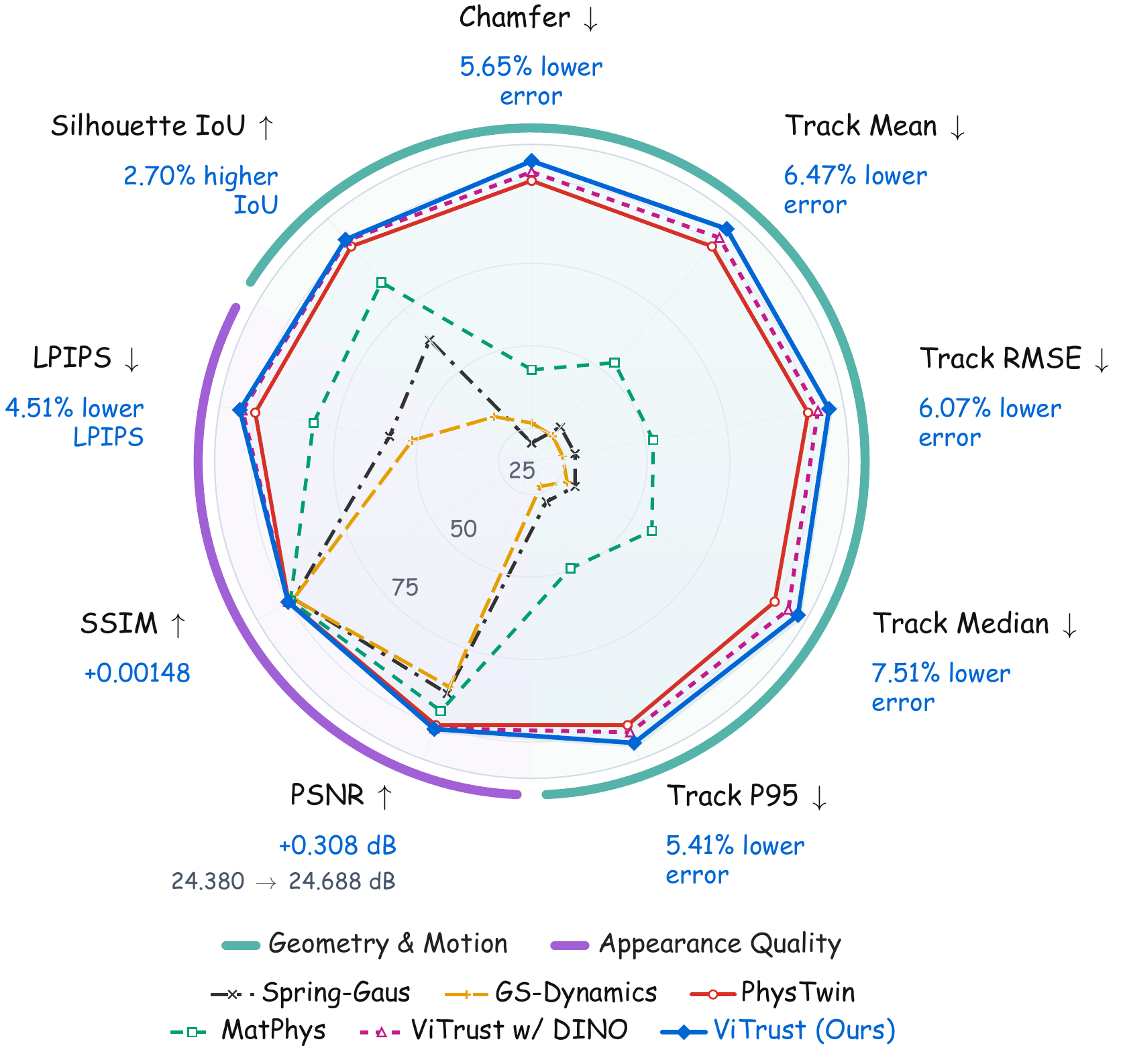}
  \caption{\textbf{Consistent gains in future physical prediction.}
  Scene-equal averages over 22 sequences and 735 future frames,
  normalized to PhysTwin (100); outward is better.
  Outer bands group geometry and motion (teal) and appearance quality (purple).
  Blue labels show gains over PhysTwin.}
  \label{fig:macro_improvement}
\end{figure}

On 22 real interaction sequences and 735 future frames, \method improves
geometry, tracking, and rendering metrics under the same simulator,
initialization, and final physical fitting budget as PhysTwin
(Figure~\ref{fig:macro_improvement}), reducing future Chamfer distance and
mean tracking error by 5.65\% and 6.47\%, respectively.
Ablations examine the roles of visual evidence, contact context, weight
regularization, and interaction history. Novel-control and cross-object
simulations additionally illustrate uses of the fitted physics.

\textbf{Our contributions are threefold:}
\begin{itemize}[leftmargin=*,nosep]
  \item We introduce \method, a framework for learning 3D observation
  trust in deformable physical identification. It jointly learns
  pointwise, time-varying observation weights and physical parameters
  through a regularized, shared-forward reconstruction objective.

  \item We develop a visual--contact trust adapter that combines
  pointwise spatiotemporal evidence with object--object and
  hand--object interaction context. Persistent patch assignments
  and recurrent temporal modeling preserve local interaction
  histories to inform observation weighting.

  \item We demonstrate improved future prediction on 22 real-world
  interaction sequences. Under the same simulator and physical
  initialization as PhysTwin, \method reduces scene-averaged future
  Chamfer distance, tracking error, and LPIPS by 5.65\%, 6.47\%,
  and 4.51\%, respectively.
\end{itemize}

\section{Related Work}

\noindent\textbf{Physical identification from videos.}
Static 3D Gaussian Splatting~\cite{kerbl2023gaussians} reconstructs scene
appearance.
D-NeRF~\cite{pumarola2021dnerf} and dynamic Gaussian representations
~\cite{luiten2024dynamic,wu2024fourDgs,yang2024deformablegs}
recover observed motion and appearance; physics-aware methods incorporate
simulation and dynamics priors
~\cite{xie2024physgaussian,zhang2024physdreamer,li2023pacnerf}.
Spring-Gaus~\cite{zhong2024springgaus} couples 3D Gaussians with a
differentiable spring--mass model, separating appearance and geometry
reconstruction from physical fitting to predict motion under changed
initial states and environments.
PhysTwin~\cite{jiang2025phystwin} combines generative geometry completion,
spring--mass dynamics, and Gaussian rendering, fitting dense physical
parameters to observed geometry and motion in sparse interaction videos.
Its tracking loss masks invalid points but assigns a common coefficient
to valid observations. \method retains PhysTwin's simulator and physical
parameterization while learning how strongly each observation constrains
the fit.

\noindent\textbf{Learning physical dynamics.}
Building on graph-based simulation~\cite{sanchez2020learning,pfaff2021learning},
GS-Dynamics~\cite{zhang2024gsdynamics} tracks dense 3D Gaussians from
multiview interaction videos and trains a graph neural dynamics model on
downsampled control particles. Predicted particle motion drives Gaussian
rendering, enabling action-conditioned prediction and manipulation planning.
MatPhys~\cite{yang2026matphys} instead predicts spring--mass parameters
from a single-view video. DINO-based part decomposition, part-level material
priors, and a shared material codebook encourage consistent physical
parameters across objects and interactions.
DeformMaster~\cite{li2026deformmaster} augments differentiable Material
Point Method simulation with particle--grid neural residuals, models
hand interactions through distributed compliant actuators, and represents
heterogeneous materials with a spatial mixture of constitutive experts.
PhysReal~\cite{deng2026physreal} combines analytical constitutive experts
with neural residuals, while PhysCoRe and PGRD
~\cite{yin2026physcore,patel2026pgrd} learn corrections to simulated dynamics.
These approaches address limited dynamics expressiveness, material
ambiguity, and simulator mismatch. Their effectiveness nevertheless depends
on the visual evidence used for fitting or supervision: tracking errors
can bias physical estimates, while accurately tracked regions with little
deformation may weakly constrain stiffness. Improving the dynamics model
alone does not resolve this uneven supervision. \method therefore learns
pointwise, time-varying fitting weights from visual and contact evidence,
and couples their learning to physical reconstruction through a regularized,
shared-forward objective. The learned component controls the relative influence of
observations rather than adding a dynamics correction.

\section{Preliminary: Spring--Mass Model}
\label{sec:preliminaries}

\noindent\textbf{Spring--mass system.}
We adopt the PhysTwin simulator~\cite{jiang2025phystwin}, which represents
an object as a graph $\mathcal G_0=(\mathcal V,\mathcal E_{\rm spr})$
of mass nodes connected by springs. Node $i$ has mass $M_i$, position
$\vect x_i$, and velocity $\vect v_i$. For spring $s=(i,j)$, let
$l_s^0$ and $k_s$ denote its rest length and stiffness, and define
$l_{ij}=\|\vect x_j-\vect x_i\|_2$ and
$\widehat{\vect d}_{ij}=(\vect x_j-\vect x_i)/l_{ij}$.
The force on node $i$ is
\begin{equation}
 \vect f_{ij}^{\rm spr}=
 \left[k_s\left(\frac{l_{ij}}{l_s^0}-1\right)
 +c_d(\vect v_j-\vect v_i)^\top\widehat{\vect d}_{ij}\right]
 \widehat{\vect d}_{ij}.
 \label{eq:spring_force}
\end{equation}
The elastic term resists stretching and compression, with $k_s$ controlling
the resistance to deformation; the damping term, with coefficient $c_d$,
opposes relative motion along the spring. Node $j$ receives the opposite
force.

Measured hand-control points act through controller links, and spring
forces propagate this local actuation through the object. The simulator
updates node velocities and positions from the accumulated forces and
gravity, with damping and contact response applied at each substep.
Restitution and friction govern rebound and sliding at contacts.
Composing these substeps gives the frame-level transition
\begin{equation}
 \vect X_{t+1}=F_{\theta,\mathcal G_0}(\vect X_t,\vect c_t,\vect c_{t+1}),
 \label{eq:state_transition}
\end{equation}
where $\vect X_t$ contains node positions and velocities, and $\vect c_t$
contains controller positions. We identify an independent stiffness $k_s$
for each spring and four shared collision scalars: ground and object--object
restitution and friction. Together these form $\theta$; topology, masses,
rest lengths, controller links, and damping remain fixed. Simulator details
are given in Appendix~\ref{app:stage2}.

\noindent\textbf{Coupling with 3D Gaussian splatting.}
A canonical 3D Gaussian representation $\mathcal A_0$ captures the object's
appearance, while the spring graph supplies its motion. Each Gaussian has
a center, orientation, scale, opacity, and appearance coefficients, and
inherits motion from the spring--mass system without separate physical
parameters of its own. It uses 16 nearby simulation nodes with normalized inverse-distance weights
recomputed at each frame. Node displacements and local
rotations inferred from neighboring nodes are transferred to Gaussian
centers and orientations through a skinning map $\mathcal B$:
\begin{equation}
 \mathcal A_{t+1}=\mathcal B(\mathcal A_t,\vect X_t,\vect X_{t+1}),
 \qquad \hat I_t^\nu=\mathcal R_\nu(\mathcal A_t).
 \label{eq:physics_rendering}
\end{equation}
Here, $\mathcal R_\nu$ renders the deformed Gaussians from camera $\nu$;
scales, opacities, and appearance coefficients remain fixed. This coupling
turns the physical rollout into rendered video, so the identified
parameters govern both predicted motion and its visual appearance.
The skinning rule is not learned; its geometric weights are distinct from
our observation weights. Appendix~\ref{app:gaussian_skinning} gives the update.

\section{Method}
\label{sec:method}

\begin{figure*}[t]
  \centering
  \includegraphics[width=\textwidth]{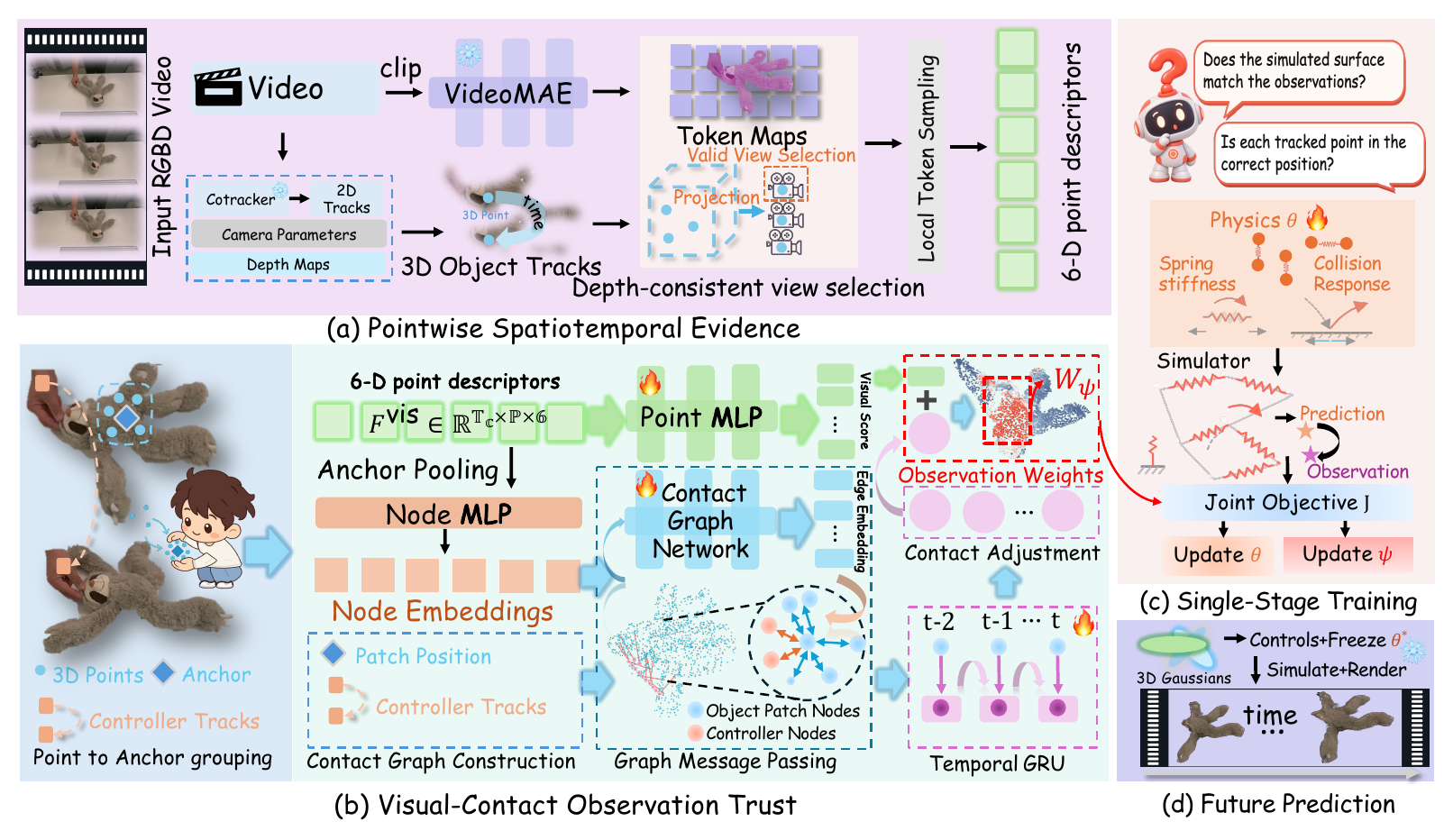}
  \caption{\textbf{Overview of \method.}
  \textbf{(a) Pointwise Spatiotemporal Evidence.}
  Depth-consistent sampling of frozen VideoMAE features, combined
  with depth and motion cues, forms pointwise spatiotemporal descriptors.
  \textbf{(b) Visual--Contact Observation Trust.}
  Contact graph message passing and a temporal GRU encode interaction
  history on persistent patches defined by fixed point-to-anchor
  assignments.
  Patch-level contact adjustments are broadcast back to points and
  fused with pointwise visual scores to predict relative observation
  weights $\vect W_\psi$ for physical identification.
  \textbf{(c) Single-Stage Training.}
  A regularized objective jointly updates physical parameters $\theta$
  and adapter parameters $\psi$ through trust-weighted geometry and
  tracking losses from a shared forward simulation on the observed prefix.
  \textbf{(d) Future Prediction.}
  Frozen physical parameters predict motion under future controls
  without refitting or future object observations, with 3D Gaussians
  used for rendering.
  Flame and snowflake symbols indicate optimized and frozen
  components, respectively.}
  \label{fig:pipeline}
\end{figure*}

\method learns observation influence by combining visual--contact evidence
with differentiable physical reconstruction. Sections~\ref{sec:videomae_evidence}
and~\ref{sec:graph} define the evidence and trust adapter;
Sec.~\ref{sec:joint_training} couples its learning to physical identification.
Algorithm~\ref{alg:v2_videomae} in Appendix~\ref{app:algorithm} summarizes
the complete pipeline, from observation preprocessing to single-stage
training and future prediction.

\subsection{Problem Formulation}

We observe a calibration prefix $\mathcal S=\{0,\ldots,T_c-1\}$ of a
$T$-frame interaction. Its observations $\mathcal O_{\mathcal S}$ comprise
calibrated multiview RGB-D images $(I_t^\nu,D_t^\nu)$, tracked 3D points
$\vect y_{t,p}\in\mathbb R^3$, visibility masks $m_{t,p}\in\{0,1\}$,
and controller positions $\vect c_t$. Indices $\nu,t,p$ denote cameras,
frames, and the $P$ tracked points. We seek parameters $\theta^*$ of the
simulator in Sec.~\ref{sec:preliminaries} that predict motion on future
frames $\mathcal Q=\{T_c,\ldots,T-1\}$ under subsequent controller inputs.
Only calibration observations enter learning and fitting; future object
observations are reserved for evaluation.

Visible measurements vary in both consistency and information about the
dynamics. We model their relative influence with a scene-specific
visual--contact adapter $f_\psi$ that predicts a weight for each point
observation $(t,p)$:
\begin{equation}
 \vect W_\psi=f_\psi(\mathcal O_{\mathcal S})
 =[w_{t,p}(\psi)]\in\mathbb R^{T_c\times P}.
 \label{eq:observation_model}
\end{equation}
The learned loss weight $w_{t,p}$ represents \emph{observation trust},
not an estimated inverse noise variance or correctness probability.
The parameters $\psi$ comprise the Point MLP, contact graph network, and
temporal GRU; VideoMAE remains frozen. We jointly optimize the physical
parameters and observation adapter using a regularized reconstruction
criterion over the observed prefix:
\begin{equation}
 \mathcal J(\theta,\psi;\mathcal S)=
 \frac{1}{|\mathcal S^+|}\sum_{t\in\mathcal S^+}
 \mathcal J_t(\theta,\psi),
 \label{eq:precision_problem}
\end{equation}
where $\mathcal S^+=\mathcal S\setminus\{0\}$ excludes the initialization
frame. Section~\ref{sec:joint_training} defines $\mathcal J_t$ and its
frame-local partial gradients and the calibration-loss criterion for
checkpoint selection.

\subsection{Pointwise Spatiotemporal Evidence}
\label{sec:videomae_evidence}

\noindent\textbf{Local video evidence.}
Frozen CoTracker3~\cite{karaev2024cotracker3} provides 2D tracks that are
lifted to 3D using depth and camera calibration.
We reproject each 3D observation into the calibrated views, select the view
with the smallest valid point-to-depth disagreement, and bilinearly sample
its frozen VideoMAE-Base token map~\cite{tong2022videomae}. Let
$\vect h^{(c)}_{t,p}$ be its unit-normalized feature from clip $c$, and
$\vect h^{\rm ref}_p$ the same track's frame-zero reference from the first
clip. Appearance agreement is measured by
\begin{equation}
 s^{\rm VM}_{t,p}=\frac{1}{|\mathcal C(t)|}
 \sum_{c\in\mathcal C(t)}
 (\vect h^{(c)}_{t,p})^\top\vect h^{\rm ref}_p.
 \label{eq:videomae_cosine}
\end{equation}
Here, $\mathcal C(t)$ contains clips covering frame $t$. Temporally
contextualized features are compared with the initial reference, and all
clips remain within $\mathcal S$ (Appendix~\ref{app:video_tokens}).

\noindent\textbf{Geometry and trajectory evidence.}
To complement appearance, we combine rescaled similarity and depth
consistency through their geometric mean $\gamma^{\rm vd}_{t,p}$.
The selected-view depth residual
$d^{\rm dep}_{t,p}$ and binary acceptance cue $\chi_{t,p}$ from confidence
screening and Kalman innovation tests provide complementary reliability cues.
With motion and time, the six-dimensional point descriptor is
\begin{equation}
 \vect f^{\rm vis}_{t,p}=[\gamma^{\rm vd}_{t,p},s^{\rm VM}_{t,p},
 \rho_d(d^{\rm dep}_{t,p}),\chi_{t,p},d^{\rm mov}_{t,p},t/(T_c-1)].
 \label{eq:visual_descriptor}
\end{equation}
Here, $\rho_d$ is a fixed logarithmic depth compression, and
$d^{\rm mov}_{t,p}=\|\vect y_{t,p}-\vect y_{t-1,p}\|_2$ for $t>0$
and zero at frame zero. Time is normalized by the observed-prefix length;
all observations are from the prefix. Fitting retains the original tracks
and masks. Standardizing channels over visible dynamic calibration samples
gives $\widetilde{\vect f}^{\rm vis}_{t,p}$. Appendix~\ref{app:local_diagnostics}
specifies numerical transforms and screening thresholds.

\subsection{Visual--Contact Observation Trust}
\label{sec:graph}

The adapter couples pointwise evidence with recurrent patch-level
interaction context through a fixed assignment (Fig.~\ref{fig:pipeline}b).

\noindent\textbf{Persistent object patches.}
Farthest-point sampling on the initial tracked positions selects anchors
$\{\vect a_i\}_{i=1}^{N_o}$. We fix point-to-patch membership by
\begin{equation}
 a(p)=\arg\min_i\|\vect y_{0,p}-\vect a_i\|_2^2,\quad
 A_{pi}=\mathbf1[a(p)=i].
 \label{eq:patch_assignment}
\end{equation}
Here, $\vect A\in\{0,1\}^{P\times N_o}$. For each nonempty patch
$\mathcal P_i=\{p:a(p)=i\}$, its position and
visibility-weighted descriptor are
\begin{equation}
 \vect o_{t,i}=\frac1{|\mathcal P_i|}
 \sum_{p\in\mathcal P_i}\vect y_{t,p},\quad
 \vect f_{t,i}=\frac{\sum_{p\in\mathcal P_i}m_{t,p}\vect f^{\rm vis}_{t,p}}
 {\sum_{p\in\mathcal P_i}m_{t,p}}.
 \label{eq:patch_pool_main}
\end{equation}
The descriptor averages all assigned points when none are visible.
A shared node encoder gives
$\vect h^{\rm node}_{t,i}=\phi_{\rm node}(\widetilde{\vect f}_{t,i})$.
Graph positions (\emph{Anchor Position} in the figure) follow patch centroids,
not individual anchor tracks. Membership stays fixed; point, node, and edge
features use separate calibration statistics. The matrix implementation is
in Appendix~\ref{app:persistent_patches}.

\noindent\textbf{Contact context over time.}
Let $\mathcal N_i^{\rm oo}$ contain the $k_o$ nearest patch neighbors at
initialization, and $\mathcal N_{t,i}^{\rm ho}$ the $k_h$ nearest controller
samples at frame $t$. Object edges $\vect e^{\rm oo}_{t,ij}$ encode
relative position and distance; contact edges $\vect e^{\rm ho}_{t,ij}$
encode separation, relative motion, surface alignment, observation
consistency, and hand identity. Relation-specific maps form messages
\begin{equation}
 \begin{aligned}
 &\vect m^{\rm oo}_{t,ij}
 =\phi_{\rm oo}([\vect h^{\rm node}_{t,i},
 \vect h^{\rm node}_{t,j},\widetilde{\vect e}^{\rm oo}_{t,ij}]),\\
 &\vect m^{\rm ho}_{t,ij}
 =\phi_{\rm ho}([\vect h^{\rm node}_{t,i},
 \widetilde{\vect e}^{\rm ho}_{t,ij}]).
 \end{aligned}
 \label{eq:relation_messages}
\end{equation}
The contact map conditions its messages on a four-channel soft latent
state and a nonnegative conditioning scalar. These heads are learned through
the observation weights, without explicit contact-state supervision in
the joint objective.
With $\operatorname{Agg}$ concatenating neighbor-wise mean and elementwise
maximum, a shared projection $\phi_a$ combines both relations:
\begin{equation}
 \begin{aligned}
 &\vect u_{t,i}=\phi_a\bigl([\vect h^{\rm node}_{t,i},
 \operatorname*{Agg}_{j\in\mathcal N_i^{\rm oo}}\vect m^{\rm oo}_{t,ij},
 \operatorname*{Agg}_{j\in\mathcal N_{t,i}^{\rm ho}}\vect m^{\rm ho}_{t,ij}]\bigr),\\
 &\vect r_{t,i}=\operatorname{GRU}_{\psi}(\vect u_{t,i},\vect r_{t-1,i}).
 \end{aligned}
 \label{eq:graph_branch}
\end{equation}
Starting at $\vect r_{-1,i}=\vect0$, the state retains patch history as
controller neighbors change. Edge descriptors and network layers are in
Appendices~\ref{app:graph} and~\ref{app:architecture}.

\noindent\textbf{Observation Weights.}
The Point MLP $\phi_{\rm vis}$ and linear contact readout $\phi_{\rm con}$
produce pointwise and patch-level scores, respectively:
\begin{equation}
 \alpha^{\rm vis}_{t,p}=\phi_{\rm vis}(\widetilde{\vect f}^{\rm vis}_{t,p}),
 \quad
 \alpha^{\rm con}_{t,i}=\phi_{\rm con}(\vect r_{t,i}).
 \label{eq:point_branch}
\end{equation}
Broadcasting contact context to the assigned points gives
\begin{equation}
 z_{t,p}=\alpha^{\rm vis}_{t,p}+\alpha^{\rm con}_{t,a(p)},\quad
 \ell_{t,p}=\ell_{\max}\tanh(z_{t,p}),
 \label{eq:trust_decomposition}
\end{equation}
where $\ell_{\max}>0$ bounds the scores. Points share contact history while
retaining their visual distinctions. For the visible set
$\mathcal V_t=\{p:m_{t,p}=1\}$ with $n_t=|\mathcal V_t|>0$, we form
bounded relative weights
\begin{equation}
 w_{t,p}=m_{t,p}\operatorname{clip}_{[0.5,2]}
                 (a_t e^{\ell_{t,p}}),\quad
 \sum_{p\in\mathcal V_t}w_{t,p}=n_t.
 \label{eq:relative_weight}
\end{equation}
The positive scalar $a_t$ is solved by bisection, so the visible mean
remains one even when some weights saturate. This preserves relative
influence while bounding both suppression and amplification. Invisible
points receive zero weight. Appendix~\ref{app:relative_precision} gives
the implicit gradient of the constrained normalization.

\subsection{Single-Stage Joint Optimization}
\label{sec:joint_training}
\label{sec:physical_refinement}

We jointly identify physical parameters $\theta$ and learn the observation
adapter $\psi$ from the observed prefix. Visual--contact evidence determines
how strongly each observation influences physical identification, while
reconstruction provides the learning signal.

\noindent\textbf{Trust-weighted reconstruction.}
Let $\hat{\vect x}_{t,j}(\theta)$ denote simulated positions,
$\mathcal X_t$ the surface-node indices, and $\mathcal M_t$ the valid
track correspondences. Each tracked point is associated with a simulation
node at initialization. We combine surface alignment and correspondence
consistency through
\begin{equation}
 \begin{aligned}
 &\mathcal L_{\rm ch}^{w}(t)=\frac1{|\mathcal V_t|}
 \sum_{p\in\mathcal V_t}w_{t,p}\min_{j\in\mathcal X_t}
 \|\hat{\vect x}_{t,j}-\vect y_{t,p}\|_2^2,\\
 &\mathcal L_{\rm tr}^{w}(t)=\frac1{3|\mathcal M_t|}
 \sum_{p\in\mathcal M_t}w^{\rm mot}_{t,p}
 \sum_{d=1}^{3}\rho(\hat x_{t,p,d}-y_{t,p,d}).
 \end{aligned}
 \label{eq:joint_reconstruction}
\end{equation}
Here, $\rho$ is the scalar SmoothL1 penalty and
$w^{\rm mot}_{t,p}=\min(w_{t-1,p},w_{t,p})$ requires trust at both ends of
a correspondence. The reconstruction loss is
$\mathcal L_{{\rm rec},t}=\mathcal L_{\rm ch}^{w}(t)+\mathcal L_{\rm tr}^{w}(t)$.
Both terms normalize by valid-point counts independently of the weights.

\noindent\textbf{Regularized joint optimization.}
Learning weights from reconstruction alone can favor suppressing
high-error observations. The unit mean and positive bounds in
Eq.~\eqref{eq:relative_weight} prevent uniform suppression and retain a
contribution from every visible point. We additionally regularize the
distribution and temporal variation of influence:
\begin{equation}
 \begin{aligned}
 &\mathcal J_t(\theta,\psi)=\mathcal L_{{\rm rec},t}(\theta,\vect W_\psi)
                            +s_tR_t(\vect W_\psi),\\
 &R_t=\lambda_{\rm KL}R_{{\rm KL},t}
      +\lambda_{\rm temp}R_{{\rm temp},t}
      +\lambda_{\rm ESS}R_{{\rm ESS},t}.
 \end{aligned}
 \label{eq:joint_objective}
\end{equation}
The regularizers encourage near-uniform weights, temporal consistency, and
sufficient effective sample support; their strength follows the unweighted
reconstruction error through the detached scale $s_t$.
Each frame simulation supplies both partial gradients of $\mathcal J_t$,
holding weights fixed for physics and simulated residuals fixed for the
adapter. Both parameter sets are updated once from the same pre-update
prediction, sharing one forward simulation.
We select the saved checkpoint with the lowest mean weighted reconstruction
loss $\mathcal L_{{\rm rec},t}$ over the calibration prefix, evaluated
during the corresponding training epoch. This score excludes regularizers
and uses no future object observations or test metrics.
Optimization, regularization, and selection details are given in
Appendices~\ref{app:joint_implementation} and~\ref{app:observation_loss}.

\noindent\textbf{Future prediction.}
After joint identification, we freeze the selected physical parameters and
simulate future motion under supplied controller inputs, without further
fitting or future object observations. The adapter is used only for
identification. Predicted motion is transferred to 3D Gaussians for rendering
through
Eq.~\eqref{eq:physics_rendering}.

\begingroup
\setlength{\dbltextfloatsep}{12pt plus 2pt minus 2pt}
\setlength{\textfloatsep}{12pt plus 2pt minus 2pt}
\setlength{\dblfloatsep}{10pt plus 2pt minus 2pt}
\section{Experiments and Results}
\label{sec:experiments}

\begin{figure*}[!t]
  \centering
  \includegraphics[width=\textwidth,height=0.79\textheight,keepaspectratio]{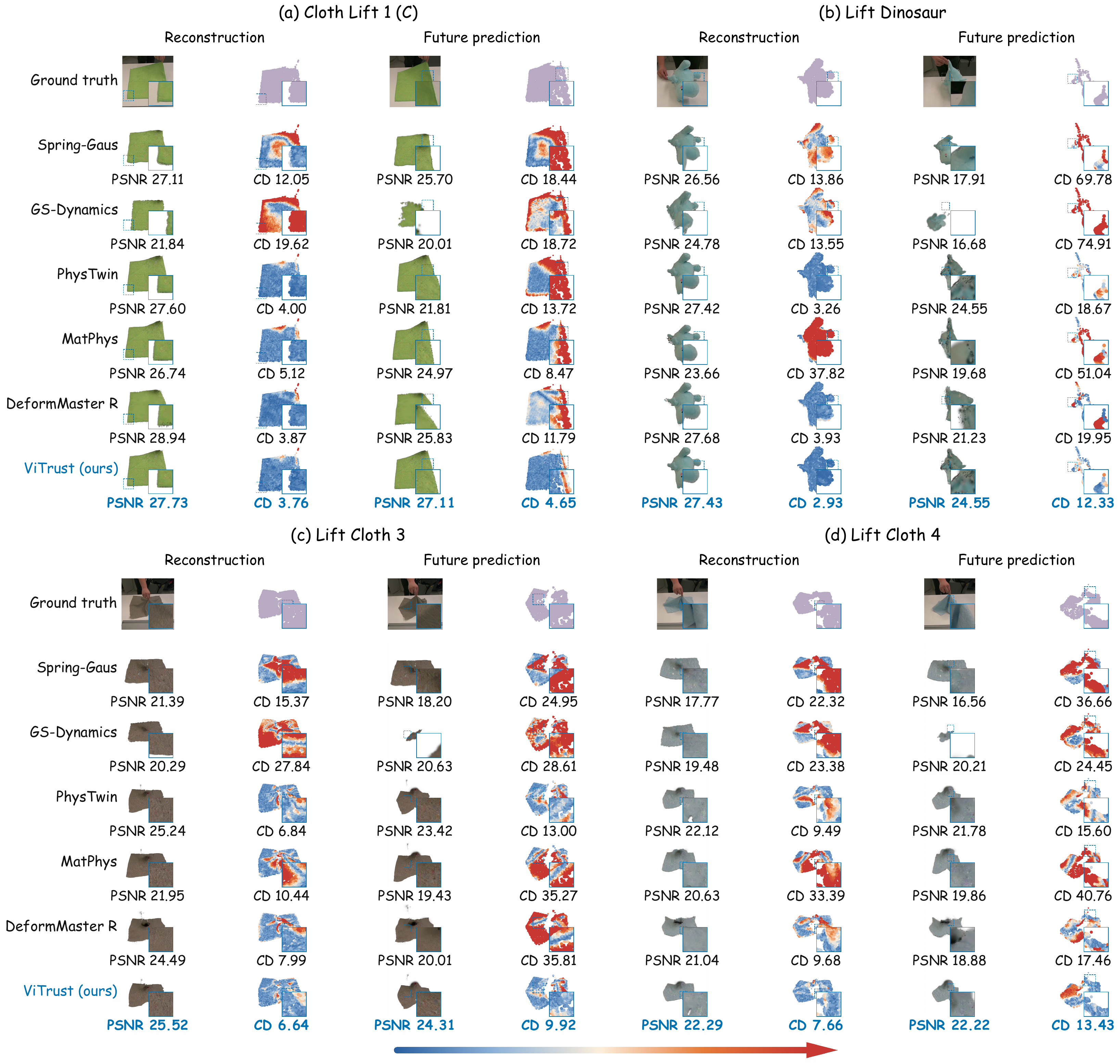}
  \caption{\textbf{Focused reconstruction and future-prediction comparison.}
  We select four interactions with large future-frame CD reductions.
  Each interaction pairs calibration reconstruction (left) with future prediction
  (right). RGB renders and geometric errors use the same frame and predicted
  trajectory, with shared crops and local zoom-in regions across methods.
  Error maps show one-sided
  L1 distances from common visible GT points to each predicted surface, using
  a shared color scale. The arrow indicates increasing pointwise error from
  blue (0 mm) to red ($\geq 30$ mm). Colors saturate at 30 mm; labels report
  full-frame PSNR (dB) and mean CD (mm), not inset-only metrics.
  Mean CD is computed from unclipped distances.
  MatPhys uses the public-checkpoint
  configuration in Appendix~\ref{app:evaluation}. DeformMaster (NeurIPS 2026)
  uses the configuration in Appendix~\ref{app:evaluation}.}
  \label{fig:reconstruction_future_focus}
\end{figure*}

\begin{figure}[!t]
  \centering
  \includegraphics[width=0.9\columnwidth]{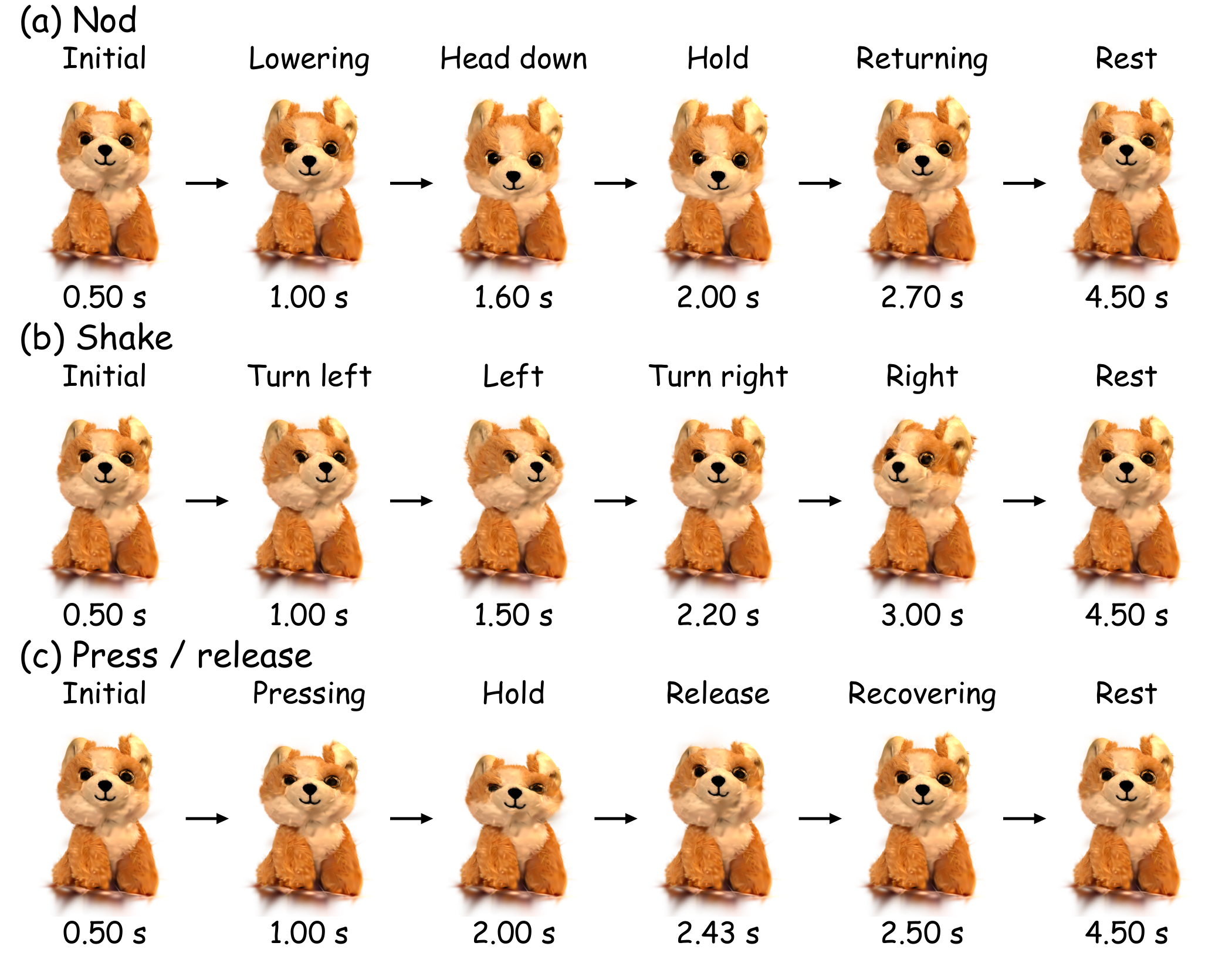}
  \caption{\textbf{Cross-object physics transfer under three scripted controls.}
  Frozen parameters identified from Lift Sloth are transferred to a static
  plush-dog reconstruction~\cite{smartdatascan2025plushdog}, without fitting
  target motion. }
  \label{fig:dog_action_sequences}
\end{figure}

\providecolor{tableoneblue}{HTML}{E6F2FA}
\begin{table*}[htbp]
\centering\footnotesize\setlength{\tabcolsep}{2.6pt}
\caption{\textbf{Reconstruction/resimulation and future prediction on 22 scenes.} Scene-equal means for the calibration prefix (left) and 735 held-out future frames (right); tracking uses available manual annotations. Distances are in mm; PSNR in dB; SSIM, LPIPS, and IoU are scaled by $100$. Baseline-specific settings are in Appendix~\ref{app:evaluation}. Best and second-best numerical values across displayed rows are bold and underlined, using unrounded values.}
\label{tab:full22_all}
\resizebox{\textwidth}{!}{%
\begin{tabular}{lcccccc@{\hspace{7pt}}cccccc}
\toprule
& \multicolumn{6}{c}{\cellcolor{reconstructiongreen}Reconstruction \& Resimulation} & \multicolumn{6}{c}{\cellcolor{predictionpink}Future Prediction}\\
\cmidrule(lr){2-7}\cmidrule(lr){8-13}
Method & CD $\downarrow$ & Track $\downarrow$ & IoU $\uparrow$ & PSNR $\uparrow$ & SSIM $\uparrow$ & LPIPS $\downarrow$ & CD $\downarrow$ & Track $\downarrow$ & IoU $\uparrow$ & PSNR $\uparrow$ & SSIM $\uparrow$ & LPIPS $\downarrow$\\
\midrule
Spring-Gaus (ECCV 2024) & 31.24 & 37.36 & 64.99 & 25.38 & 93.91 & 6.50 & 63.75 & 92.40 & 38.94 & 21.89 & 91.96 & 11.12\\
GS-Dynamics (CoRL 2024) & 39.65 & 56.06 & 49.95 & 22.77 & 92.78 & 9.64 & 49.23 & 105.66 & 20.18 & 21.38 & 92.07 & 12.63\\
PhysTwin (ICCV 2025) & \underline{5.59} & \best{9.36} & 74.91 & 26.27 & 94.35 & 3.98 & \underline{13.11} & \underline{26.40} & \underline{61.79} & \underline{24.38} & \underline{93.66} & \underline{6.53}\\
\midrule
MatPhys (arXiv 2026) & 16.02 & 25.14 & 68.99 & 25.50 & 94.04 & 5.01 & 30.68 & 48.86 & 53.01 & 23.26 & 93.07 & 7.95\\
DeformMaster (NeurIPS 2026) & 6.12 & 11.36 & \best{80.86} & \best{29.33} & \best{95.87} & \best{3.41} & 19.95 & 38.49 & 57.65 & 24.14 & 93.52 & 8.33\\
\rowcolor{tableoneblue}\textbf{ViTrust (ours)} & \best{5.51} & \underline{9.80} & \underline{75.15} & \underline{26.30} & \underline{94.36} & \underline{3.98} & \best{12.37} & \best{24.70} & \best{63.46} & \best{24.69} & \best{93.81} & \best{6.23}\\
\bottomrule
\end{tabular}%
}
\end{table*}

\newcommand{\heuristicAblationRows}{%
\midrule
CoTracker confidence & 5.49 & 9.74 & 75.24 & 26.30 & 94.37 & 3.95 & 12.91 & 25.32 & 62.46 & 24.48 & 93.73 & 6.37\\
Residual weighting & 5.64 & 10.16 & 75.02 & 26.26 & 94.34 & 3.99 & 13.11 & 25.52 & 62.73 & 24.53 & 93.74 & 6.35\\
}

\begin{table*}[htbp]
\centering\footnotesize\setlength{\tabcolsep}{2.6pt}
\setlength{\aboverulesep}{0.75pt}\setlength{\belowrulesep}{1pt}
\renewcommand{\arraystretch}{0.97}
\caption{\textbf{Observation-weighting ablations.} Scene-equal means over 22 scenes; units follow Table~\ref{tab:full22_all}. Bold ranks learned-weight variants. Heuristics follow Appendix~\ref{app:heuristic_weighting}; per-scene results are in Tables~\ref{tab:per_scene_ablation_reconstruction} and~\ref{tab:per_scene_ablation_prediction}.}
\label{tab:vitrust_ablation}
\resizebox{\textwidth}{!}{%
\begin{tabular}{lcccccc@{\hspace{7pt}}cccccc}
\toprule
& \multicolumn{6}{c}{\cellcolor{reconstructiongreen}Reconstruction \& Resimulation} & \multicolumn{6}{c}{\cellcolor{predictionpink}Future Prediction}\\
\cmidrule(lr){2-7}\cmidrule(lr){8-13}
\makebox[94pt][l]{Variant} & CD $\downarrow$ & Track $\downarrow$ & IoU $\uparrow$ & PSNR $\uparrow$ & SSIM $\uparrow$ & LPIPS $\downarrow$ & CD $\downarrow$ & Track $\downarrow$ & IoU $\uparrow$ & PSNR $\uparrow$ & SSIM $\uparrow$ & LPIPS $\downarrow$\\
\midrule
Visual trust only & 5.72 & 10.36 & \best{75.28} & \best{26.46} & \best{94.39} & 3.95 & 13.06 & 26.72 & 62.53 & 24.51 & 93.72 & 6.43\\
Contact-graph trust only & 5.54 & 9.88 & 75.11 & 26.29 & 94.36 & 3.98 & 12.77 & 25.35 & 63.37 & 24.69 & 93.79 & 6.24\\
Frame-wise (no GRU) & \best{5.51} & \best{9.73} & 75.28 & 26.34 & 94.37 & \best{3.94} & 12.83 & 25.07 & 62.74 & 24.56 & 93.78 & 6.28\\
w/o Weight regularization & 5.63 & 10.07 & 74.97 & 26.24 & 94.32 & 4.03 & 13.08 & 26.10 & 62.14 & 24.40 & 93.66 & 6.51\\
\heuristicAblationRows
\midrule
w/ DINO & 5.56 & 9.74 & 75.12 & 26.30 & 94.36 & 3.99 & 12.77 & 25.52 & 63.16 & 24.64 & 93.78 & 6.28\\
\rowcolor{tableoneblue}\textbf{ViTrust (ours)} & 5.51 & 9.80 & 75.15 & 26.30 & 94.36 & 3.98 & \best{12.37} & \best{24.70} & \best{63.46} & \best{24.69} & \best{93.81} & \best{6.23}\\
\bottomrule
\end{tabular}%
}
\end{table*}

\subsection{Experimental Setup}

\textbf{Datasets.} We conduct training and quantitative evaluation on the
PhysTwin dataset~\cite{jiang2025phystwin}, which comprises 22 real RGB-D
interaction sequences: eight plush, eight cloth, five rope, and one package
scene. The sequences capture lifting, pushing, and stretching with one or
two hands, recorded from three calibrated views at 30 FPS. We strictly follow
PhysTwin's official temporal train/test splits, using the first 70\% of
each sequence for per-scene training and the remaining 30\% for future
prediction. This yields 1,688 training frames and 735 test frames across
the 22 sequences. All fitting and checkpoint selection use only the training
prefix; test rollouts are conditioned on measured future controller motion.

\textbf{Baselines.} We compare with Spring-Gaus~\cite{zhong2024springgaus},
GS-Dynamics~\cite{zhang2024gsdynamics}, PhysTwin~\cite{jiang2025phystwin},
MatPhys~\cite{yang2026matphys}, and DeformMaster~\cite{li2026deformmaster}
using common splits and evaluation. PhysTwin and \method share the physical
initialization, simulator, and 200-epoch fitting budget. MatPhys uses a
released checkpoint; DeformMaster is a local paper-guided reproduction.
Baseline-specific settings are given in Appendix~\ref{app:evaluation}.

\textbf{Metrics.} One-sided L1 Chamfer distance (CD) measures distance from
visible ground-truth points to the simulated surface. Independent point
correspondences provide mean tracking error, RMSE, median error, and P95.
PSNR, SSIM, LPIPS~\cite{zhang2018lpips}, and silhouette IoU measure camera-0
rendering after hand pixels are excluded. All metrics use the same saved
simulation trajectory. We average frames within each scene, then average the
22 scenes equally. Calibration reconstruction and future prediction are
reported separately.

\textbf{Implementation details.} We use a frozen VideoMAE-Base encoder and
train for 200 epochs: two physics-only warm-up epochs followed by 198
joint-update epochs.
Physics uses Adam (learning rate $10^{-3}$) and the adapter AdamW
($10^{-4}$), with adapter gradient norms clipped to 5.
We train on NVIDIA H20 GPUs (96 GiB); full settings are given in
Appendix~\ref{app:joint_implementation}.

\subsection{Results}
\label{sec:results}

\subsubsection{Future Prediction}

With physical parameters fixed after fitting, we evaluate the simulator's
response to subsequent controller motion. In Table~\ref{tab:full22_all},
\method improves all six reported future metrics over PhysTwin, reducing CD and
mean tracking error by 5.65\% and 6.47\%, respectively, while lowering
LPIPS by 4.51\%. These gains span surface geometry, material-point motion,
and rendered appearance from the same predicted trajectory.
Despite a slightly higher calibration tracking error than PhysTwin
(9.80 vs.\ 9.36 mm), \method achieves lower future tracking error
(24.70 vs.\ 26.40 mm). Under the shared simulator, initialization, and
final fitting budget, this contrast shows that closer agreement with the observed
tracks does not necessarily yield better prediction, supporting the benefit of
learned observation weighting for future prediction.

The largest category gains occur on cloth, with CD and tracking error
reduced by 10.10\% and 12.61\% relative to PhysTwin; improvements also
extend to rope and plush objects.
Figure~\ref{fig:reconstruction_future_focus}
provides a spatial view of these differences: the selected future-frame
crops show better preservation of cloth folds and silhouettes, and
articulated plush shape. These local comparisons complement
the aggregate errors by showing where predicted surfaces depart from the
observed geometry. Per-scene results for all nine metrics appear in
Appendix~\ref{app:complete_results}.

\subsubsection{Reconstruction and Resimulation}

On the calibration prefix, \method achieves the best CD (5.51 mm) in the
left half of Table~\ref{tab:full22_all}. DeformMaster
has stronger calibration rendering metrics, but higher future
errors; its separate appearance refinement and differing simulation settings
prevent attributing this contrast solely to dynamics identification.
\method's tracking error remains close to the best
result (9.80 vs.\ 9.36 mm). The reconstruction columns in
Figure~\ref{fig:reconstruction_future_focus} show preserved cloth boundaries
and articulated plush structure. Thus, the regularized
joint objective also supports accurate reconstruction of the observed
interaction.

\subsubsection{Ablation Study}
\label{sec:ablation}

Table~\ref{tab:vitrust_ablation} examines how observation evidence,
interaction context, temporal memory, and weight regularization affect
the learned weights.
Removing the contact
branch (\emph{Visual trust only}) or the pointwise visual branch
(\emph{Contact-graph trust only}) raises future CD from 12.37 to
13.06 and 12.77 mm, respectively. Replacing VideoMAE with frame-wise
DINOv2~\cite{oquab2024dinov2} (\emph{ViTrust w/ DINO}) raises future tracking
error from 24.70 to 25.52 mm.
\emph{Frame-wise (no GRU)} replaces recurrent memory with a frame-wise MLP;
future tracking error rises to 25.07 mm despite a closer calibration fit
(9.73 vs.\ 9.80 mm), supporting persistent interaction history.
In a separate matched single-stage comparison,
removing the KL, temporal, and ESS penalties in
Eq.~\eqref{eq:joint_objective}, while retaining the hard
weight constraints, increases future CD from 12.66 to 13.08 mm and tracking
error from 24.81 to 26.10 mm; PSNR drops from 24.64 to 24.40 dB.
These results support the soft penalties beyond bounded, unit-mean weights.

\subsubsection{Cross-Object Transfer under Novel Controls}
\label{sec:cross_object_control}

Figure~\ref{fig:dog_action_sequences} demonstrates qualitative transfer of
frozen Lift Sloth parameters to a static plush-dog reconstruction without
fitting target motion. Object and controller springs inherit their source
median stiffnesses; damping and collision coefficients remain unchanged.
The same parameter set supports nodding, shaking, and pressing, including
passive recovery after release, illustrating reuse across geometries and
controls. Each row shows six successive states:
(a)~a $22^\circ$ nod, hold, and return;
(b)~a left--right shake with $\pm25^\circ$ head-handle rotation and return;
(c)~a 30\,mm downward crown-handle displacement, hold, and release.
The crown-control springs are detached at
2.43\,s; subsequent recovery is passive, with no upward return command.

\FloatBarrier
\endgroup
\section{Conclusion}

We present \method, a framework that learns how tracked 3D observations
should influence deformable physical identification.
By combining spatiotemporal visual evidence with interaction histories
on a persistent contact graph, \method learns pointwise, time-varying
observation weights jointly with physical parameters.
A regularized, shared-forward objective enables this learning using
only the observed motion, while the identified physics remains fixed
for future prediction under supplied controls.
Experiments on 22 real-world interaction sequences show improved
future geometry, tracking, and rendered appearance over the evaluated
state-of-the-art baselines.
These results highlight learning observation influence as a
complementary route to better physical prediction, alongside
improvements to the dynamics model itself.

\section*{LLM Usage Statement}
We used OpenAI ChatGPT and Codex to assist with language editing, LaTeX
formatting, and code debugging. All assisted content were reviewed and verified by the authors, who take full
responsibility for the scientific claims, results, and final manuscript.

{\small
\bibliographystyle{ieeenat_fullname}
\bibliography{references}
}

\clearpage
\appendix
\raggedbottom
\section{Contact Graph Construction}
\label{app:graph}

We retain the notation of Sec.~\ref{sec:method}:
$\psi$ denotes the adapter, $\vect W_\psi$ its observation weights,
and $\theta$ the physical parameters. Indices $p$ and $i$
distinguish tracked points and contact-graph patches; $k$ indexes
frame-local optimizer updates. Simulation-node
indices in Appendix~\ref{app:stage2} refer to the separate spring--mass graph.

\subsection{Depth Confidence and Tracking Consistency}
\label{app:local_diagnostics}

The fixed video--depth confidence in Eq.~\eqref{eq:visual_descriptor}
combines selected-view depth evidence with local video similarity:
\begin{equation}
 \begin{aligned}
 &\gamma^{\rm geom}_{t,p}
 =\gamma^{\rm loc}_{t,p}
 \exp[-(d^{\rm dep}_{t,p})^2/(2\sigma_d^2)],\\
 &\gamma^{\rm vd}_{t,p}
 =\sqrt{\gamma^{\rm geom}_{t,p}\mathcal N_{\rm VM}(s^{\rm VM}_{t,p})}.
 \end{aligned}
 \label{eq:confidence_main}
\end{equation}
Here, $d^{\rm dep}_{t,p}$ is the absolute depth disagreement in the selected
view and $\gamma^{\rm loc}_{t,p}$ measures neighborhood depth consistency.
The geometric mean is high only when both sources of evidence agree.
We set $\sigma_d=0.015\,\mathrm{m}$ and use the clipped affine mapping
\begin{equation}
 \mathcal N_{\rm VM}(s)=\operatorname{clip}
 \left(\frac{s-\mu_{\rm VM}}{\Delta_{\rm VM}},0,1\right).
\end{equation}
The offset and scale are $\mu_{\rm VM}=0.25$ and $\Delta_{\rm VM}=0.65$.
The depth compression in Eq.~\eqref{eq:visual_descriptor} is
\begin{equation}
 \rho_d(d)=\log\!\left(1+\frac{\min(d,d_{\max})}{\sigma_r}\right),
\end{equation}
where $d_{\max}=0.25\,\mathrm{m}$ caps the nonnegative residual and
$\sigma_r=0.003\,\mathrm{m}$ sets its scale.

For the selected view, let $\mathcal Z_{t,p}$ be the valid depths in the
$5\times5$ pixel neighborhood, $\widetilde d_{t,p}$ their median,
$\operatorname{MAD}_{t,p}$ their median absolute deviation, and
$d^{\rm proj}_{t,p}$ the
projected point depth. The local depth confidence is
\begin{equation}
 \gamma^{\rm loc}_{t,p}=\exp\!\left(
 -\frac{|d^{\rm proj}_{t,p}-\widetilde d_{t,p}|}{0.02}\right)
 \exp\!\left(-\frac{\operatorname{MAD}_{t,p}}{0.01}\right).
 \label{eq:local_depth_confidence_app}
\end{equation}

To construct $\chi_{t,p}$ in Eq.~\eqref{eq:visual_descriptor}, we first
retain visible observations with joint confidence at least 0.15.
A constant-acceleration Kalman filter~\cite{kalman1960} then accepts an
observation when its squared Mahalanobis innovation is at most 16.27;
the first frame of each segment uses confidence screening alone.
The filter uses process noise 25 and measurement standard deviation
0.003 m, and runs independently within the official temporal segments.
Only the forward-pass acceptance cue enters the adapter; physical
identification uses the original trajectories and visibility masks,
not smoothed tracks.

\subsection{Point Evidence and Persistent Patches}
\label{app:persistent_patches}

Let $\vect Y_t\in\mathbb R^{P\times3}$ and
$\vect F_t^{\rm vis}\in\mathbb R^{P\times6}$ stack tracked positions and
descriptors. Farthest-point sampling on $\vect Y_0$ selects
$N_o=\min(P,512)$ anchors $\{\vect a_i\}_{i=1}^{N_o}$. Sampling and the assignment
in Eq.~\eqref{eq:patch_assignment} are performed only at initialization.
With the visibility matrix
$\vect M_t=\operatorname{diag}(m_{t,1},\ldots,m_{t,P})$, define
$\vect P=\operatorname{RowNorm}(\vect A^\top)$ and
$\vect P_t=\operatorname{RowNorm}(\vect A^\top\vect M_t)$.
$\operatorname{RowNorm}$ divides each nonzero row by its sum; a zero row
of $\vect P_t$ is replaced by the corresponding row of $\vect P$.
The matrix form of Eq.~\eqref{eq:patch_pool_main} is
\begin{equation}
 \vect O_t=\vect P\vect Y_t,\qquad
 \vect F_t^{\rm node}=\vect P_t\vect F_t^{\rm vis}.
 \label{eq:patch_pool_matrix_app}
\end{equation}
Their rows are $\vect o_{t,i}^\top$ and $\vect f_{t,i}^\top$,
respectively.

\subsection{Normals, Hand Identity, and Edges}

Let $\vect o_{t,i}$ and $\vect c_{t,j}$ be the positions of object patch
$i$ and controller $j$. Their displacement and relative velocity are
$\vect\delta_{t,ij}=\vect o_{t,i}-\vect c_{t,j}$ and
$\vect v^{\rm rel}_{t,ij}=\dot{\vect o}_{t,i}-\dot{\vect c}_{t,j}$, where dots denote
temporal derivatives. With object and hand unit normals $\vect n^o_{t,i}$
and $\vect n^h_{t,j}$, the hand--object edge descriptor is
\begin{equation}
 \begin{aligned}
 &\vect e^{\rm ho}_{t,ij}=[d_{t,ij},\delta^n_{t,ij},v^n_{t,ij},
 v^\tau_{t,ij},(\vect n^o_{t,i})^\top\vect n^h_{t,j},\\[-2pt]
 &\quad\gamma^{\rm cos}_{t,i},\gamma^{\rm dep}_{t,i},
 \gamma^{\rm trk}_{t,i},\operatorname{onehot}(\mathrm{hand}_j)],
 \end{aligned}
 \label{eq:edge_features}
\end{equation}
Here, $d=\|\vect\delta\|_2$ is separation distance,
$\delta^n=\vect\delta^\top\vect n^h$ is normal displacement,
$v^n=(\vect v^{\rm rel})^\top\vect n^h$ is normal relative velocity, and
$v^\tau=\|\vect v^{\rm rel}-v^n\vect n^h\|_2$ is tangential speed, with indices
$i,j,t$ omitted. The superscript $n$ denotes a normal component.
The normal inner product measures surface alignment.
$\gamma^{\rm cos}$, $\gamma^{\rm dep}$, and $\gamma^{\rm trk}$ denote
appearance similarity, depth consistency, and tracking reliability.
$\operatorname{onehot}(\mathrm{hand}_j)$ is the binary encoding of the
hand component containing controller $j$. These cues and the normal and
component estimates are defined below.

Object normals are the smallest-eigenvalue eigenvectors of the local
position covariance over the eight nearest nodes, including the node itself
(or all nodes when $N_o<8$). Signs point toward the nearest controller sample.
Controller normals use
the analogous eight-neighbour PCA. Their frame-0 signs point toward the
nearest initial object point, and later signs are chosen to agree with the
previous-frame normal, producing temporal orientation consistency.

To infer one or two hands, we construct the complete frame-0 controller
distance graph and its minimum spanning tree. We remove the longest edge
only when its length exceeds both 0.04 m and three times the median
tree-edge length, and both resulting components contain at least three
controller points.
Hand identities remain fixed throughout the sequence.

The $k_o=\min(8,N_o-1)$ nearest object neighbors are fixed at frame zero;
the $k_h=\min(4,N_h)$ nearest controller neighbors are recomputed at each
calibration frame, where $N_h=30$. The confidence channels in
Eq.~\eqref{eq:edge_features} are computed from pooled visual features as
\begin{equation}
 \begin{aligned}
 &\gamma^{\rm cos}_{t,i}=\tfrac12(\operatorname{clip}(s^{\rm VM}_{t,i},-1,1)+1),\\
 &d^{\rm dep}_{t,i}=\sigma_r[\exp(\max(f_{t,i,3},0))-1],\\
 &\gamma^{\rm dep}_{t,i}=\exp(-d^{\rm dep}_{t,i}/0.01),\\
 &\gamma^{\rm trk}_{t,i}=\operatorname{clip}(\gamma^{\rm vd}_{t,i},0,1)
 \operatorname{clip}(\chi_{t,i},0,1).
 \end{aligned}
\end{equation}
Here $s^{\rm VM}_{t,i}$, $\gamma^{\rm vd}_{t,i}$, and $\chi_{t,i}$ are
pooled by Eq.~\eqref{eq:patch_pool_main}, including its visibility fallback.
The scalar $f_{t,i,3}$ is the third channel of the pooled descriptor.
Depth confidence is reconstructed from this pooled log-depth feature;
$d^{\rm dep}_{t,i}$ is therefore not the arithmetic mean of raw residuals.
Object and controller velocities in contact edges are backward finite
differences multiplied by 30 FPS and are zero at frame zero.

\section{Feature Extraction and Tensor Shapes}
\label{app:tensor_specification}

Table~\ref{tab:tensor_contract} lists stacked tensor shapes using the
main paper's element-wise notation. Here $V=3$, $H_I,W_I$ are image height
and width, $N_{\rm clip}$ counts clips per view, and $N_s$ counts optimized
springs. Node and neighbor counts follow Appendix~\ref{app:graph}.

\begin{table}[t]
\centering
\small
\setlength{\tabcolsep}{3pt}
\caption{Calibration tensor shapes for one scene.
VideoMAE maps are listed per camera view.}
\label{tab:tensor_contract}
\begin{tabular*}{\columnwidth}{@{}l@{\extracolsep{\fill}}r@{}}
\toprule
Quantity & Shape\\
\midrule
RGB $I_t^\nu$ & $T_c\!\times\!V\!\times\!H_I\!\times\!W_I\!\times\!3$\\
Depth $D_t^\nu$ & $T_c\!\times\!V\!\times\!H_I\!\times\!W_I$\\
Tracks $\vect y_{t,p}$ & $T_c\!\times\!P\!\times\!3$\\
Masks $m_{t,p}$ & $T_c\!\times\!P$\\
Controllers $\vect c_{t,j}$ & $T_c\!\times\!30\!\times\!3$\\
VideoMAE maps $\vect H^{\rm VM}$ & $N_{\rm clip}\!\times\!8\!\times\!14\!\times\!14\!\times\!768$\\
Point descriptors $\vect f^{\rm vis}_{t,p}$ & $T_c\!\times\!P\!\times\!6$\\
Patch descriptors $\vect f_{t,i}$ & $T_c\!\times\!N_o\!\times\!6$\\
Object edges $\vect e^{\rm oo}_{t,ij}$ & $T_c\!\times\!N_o\!\times\!k_o\!\times\!4$\\
Contact edges $\vect e^{\rm ho}_{t,ij}$ & $T_c\!\times\!N_o\!\times\!k_h\!\times\!10$\\
Latent contact feature $\vect q_{t,ij}$ & $T_c\!\times\!N_o\!\times\!k_h\!\times\!4$\\
Latent contact scalar $\sigma^{\rm con}_{t,ij}$ & $T_c\!\times\!N_o\!\times\!k_h$\\
Contact score $\alpha^{\rm con}_{t,i}$ & $T_c\!\times\!N_o$\\
Visual score $\alpha^{\rm vis}_{t,p}$ & $T_c\!\times\!P$\\
Observation weights $\vect W_\psi$ & $T_c\!\times\!P$\\
Fitted physics $\theta^*$ & $N_s+4$\\
\bottomrule
\end{tabular*}
\end{table}

\subsection{RGB-D Inputs and 3D Track Preprocessing}
\label{app:track_preprocessing}

The inherited
PhysTwin~\cite{jiang2025phystwin} preprocessing uses
CoTracker3~\cite{karaev2024cotracker3} on RGB videos to establish 2D point
correspondences over time. Depth and calibration then lift tracked image
locations into a common world coordinate system using the three supplied
camera calibrations. For hand-control trajectories, the inherited pipeline segments hands with
Grounded-SAM2~\cite{ren2024grounded,ravi2024sam2} and tracks hand points
with CoTracker3. After lifting the tracks to 3D, farthest-point sampling
on the initial frame selects 30 valid control points whose identities are
retained throughout the sequence. We use these control trajectories
as inputs to both the contact graph and the simulator. The adapter and physical fitting use the same object tracks and visibility
masks.

\subsection{Local Video Tokenization and 3D Lifting}
\label{app:video_tokens}

Clips contain 16 frames with stride 8, resized to $224\times224$; their
receptive fields remain inside the calibration prefix, with last-frame
padding when needed.
The frozen VideoMAE-Base encoder uses tubelet size 2 and patch size 16,
and returns $8\cdot14\cdot14=1{,}568$ local tokens of
width 768, reshaped as
$\vect H^{\rm VM}\in\mathbb R^{8\times14\times14\times768}$.
The encoder runs independently on each camera's RGB clips. For an
unpadded frame at offset $\ell$ in a clip, we use temporal token slice
$\lfloor\ell/2\rfloor$; each slice therefore serves two consecutive RGB
frames, sampled at their respective projected point locations.
For the calibrated pixel projection $\vect u^{(\nu)}_{t,p}$ of
$\vect y_{t,p}$, the absolute depth disagreement is
$d^{{\rm dep},\nu}_{t,p}=|D_t^\nu(\vect u^{(\nu)}_{t,p})-
z^{(\nu)}_{t,p}|$, where $z^{(\nu)}_{t,p}$ is the point's camera-space depth.
We select the view with smallest valid depth disagreement and
bilinearly sample its spatial token map. Selection is per point and frame,
uses only the tracked geometry and depth, and does not depend on VideoMAE
features. Projected image coordinates are rescaled to the spatial token
grid for sampling. For each covering clip $c$, the
sampled token is normalized separately:
\begin{equation}
 \vect h^{(c)}_{t,p}=\operatorname{norm}_2\!\left[
 \operatorname{Bilinear}(\vect H^{\rm VM}_{c,t},\vect u_{t,p})\right]
 \in\mathbb R^{768}.
\label{eq:token_lifting_app}
\end{equation}
Here, $\vect H^{\rm VM}_{c,t}$ is the spatial token map for frame $t$ in
clip $c$, $\operatorname{Bilinear}$ samples it at the projected location
in the selected view, and $\operatorname{norm}_2$ normalizes the resulting
feature to unit Euclidean norm.

\subsection{Feature Normalization}

Each input channel is standardized using calibration samples only:
\begin{equation}
 \operatorname{Norm}_{d}(x_d)=\frac{x_d-\mu_d^{\rm feat}}
 {\max(s_d^{\rm feat},10^{-5})}.
\label{eq:standardization_app}
\end{equation}
Here, $d$ indexes a feature channel, $x_d$ is its value, and
$\mu_d^{\rm feat}$ and $s_d^{\rm feat}$ are its calibration mean and
population standard deviation. A tilde denotes the resulting standardized descriptor,
as in $\widetilde{\vect f}^{\rm vis}_{t,p}$ in Sec.~\ref{sec:graph}.
Point statistics use visible dynamic samples; node and edge statistics use all
calibration graph entries. Node assignments follow
Eq.~\eqref{eq:patch_assignment}, and edge construction follows
Appendix~\ref{app:graph}. A shared temporal GRU propagates each patch's
state between successive frames.

\section{Adapter Architecture and Training}
\label{app:adapter_derivation}

\subsection{End-to-End Procedure}
\label{app:algorithm}

Algorithm~\ref{alg:v2_videomae} summarizes the complete pipeline.

\begin{algorithm}[t]
\caption{Complete \method pipeline}
\label{alg:v2_videomae}
\small
\begin{algorithmic}[1]
\Require Calibrated RGB-D prefix; controls $\vect c_{0:T-1}$;
shared physical initialization $(\mathcal G_0,\theta_0,\vect X_0)$;
canonical Gaussians $\mathcal A_0$
\Ensure Parameters $(\theta^*,\psi^*)$, future states
$\hat{\vect X}_{T_c:T-1}$, and renders $\hat I_{T_c:T-1}^\nu$

\Statex \textcolor{cvprblue}{\textbf{Step 1: Pointwise Spatiotemporal Evidence}}
\State Lift tracks to 3D; sample frozen VideoMAE
tokens from depth-consistent camera views
\State Combine appearance, depth, tracking, motion,
and time cues; standardize descriptors (Eq.~\eqref{eq:visual_descriptor})

\Statex \textcolor{cvprblue}{\textbf{Step 2: Visual--Contact Observation Trust}}
\State Sample frame-zero anchors; fix point-to-patch
membership and pool node features (Eqs.~\eqref{eq:patch_assignment}--\eqref{eq:patch_pool_main})
\State Fix object--object nearest-neighbor edges;
recompute hand--object neighbors at each frame
\State Encode edge geometry, relative motion,
normals, reliability, and hand identity (Eq.~\eqref{eq:edge_features})
\State Define $f_\psi$ with relation-specific messages,
aggregation, and a temporal GRU (Eqs.~\eqref{eq:relation_messages}--\eqref{eq:graph_branch})
\State Fuse Point MLP and patch contact scores;
bound and unit-mean normalize weights (Eqs.~\eqref{eq:trust_decomposition}--\eqref{eq:relative_weight})

\Statex \textcolor{cvprblue}{\textbf{Step 3: Single-Stage Joint Optimization}}
\State Set $\theta\gets\theta_0$; initialize $\psi$ to produce
unit visible weights
\State Warm up $\theta$ for 2 epochs with unit visible weights
\State Jointly optimize $(\theta,\psi)$ for 198 epochs using
frame-local updates from a shared simulation (Eq.~\eqref{eq:joint_partials})
\State Select $(\theta^*,\psi^*)$ by minimum calibration
loss (Eq.~\eqref{eq:calibration_checkpoint_selection}); freeze $\theta^*$

\State Roll out from $(\vect X_0,\mathcal A_0)$ with
supplied controls; transfer motion to Gaussians
and render future frames (Eq.~\eqref{eq:physics_rendering})
\end{algorithmic}
\end{algorithm}

\subsection{Network Architecture}
\label{app:architecture}
\label{app:forward_pass}

The adapter $\psi$ contains 5,735 trainable parameters and uses hidden
width $H=16$. Each MLP has two affine layers with a SiLU activation
between them. The node encoder $\phi_{\rm node}$, object-message map
$\phi_{\rm oo}$, and Point MLP $\phi_{\rm vis}$ have widths
$6\!\to\!16\!\to\!16$, $36\!\to\!16\!\to\!16$, and
$6\!\to\!16\!\to\!1$, respectively. The frozen VideoMAE encoder
is not part of $\psi$.

In the expressions below, MLP subscripts give input, hidden, and output
widths. Tildes denote the standardized inputs from
Eq.~\eqref{eq:standardization_app}.
The hand--object map $\phi_{\rm ho}$ in
Eq.~\eqref{eq:relation_messages} comprises an edge encoder, latent heads,
and a contact-message MLP. Its edge embedding is
\begin{equation}
 \vect g_{t,ij}=\operatorname{MLP}_{26,16,16}
 ([\vect h^{\rm node}_{t,i},\widetilde{\vect e}^{\rm ho}_{t,ij}]).
 \label{eq:contact_embedding_app}
\end{equation}
The affine heads $H_q$ and $H_\sigma$ have dimensions
$16\!\to\!4$ and $16\!\to\!1$, mapping this embedding to four
categorical logits and one scalar logit:
\begin{equation}
 \begin{aligned}
 &\vect q_{t,ij}=\operatorname{softmax}(H_q(\vect g_{t,ij})),\\
 &\sigma^{\rm con}_{t,ij}=\operatorname{softplus}(H_\sigma(\vect g_{t,ij})).
 \end{aligned}
 \label{eq:contact_embedding_main}
\end{equation}
The four channels and nonnegative scalar condition the contact message.
They are trained through the observation weights in the single-stage
objective, without an additional state or uncertainty loss. Neither
fixed contact-state semantics nor calibrated uncertainty is assumed.
The message is
\begin{equation}
 \vect m^{\rm ho}_{t,ij}=\operatorname{MLP}_{21,16,16}
 ([\vect g_{t,ij},\vect q_{t,ij},\sigma^{\rm con}_{t,ij}]).
 \label{eq:contact_message_app}
\end{equation}
The aggregation map $\phi_a$ in Eq.~\eqref{eq:graph_branch} is an
affine--SiLU projection from $5H=80$ to $H=16$ channels. Its output
feeds a one-layer GRU with input and hidden dimensions of 16, followed
by a $16\!\to\!1$ affine contact readout. Graph construction uses
at most 512 object patches with eight object neighbors and 30 controller
nodes with four controller neighbors per patch, as specified in
Appendix~\ref{app:graph}.

\subsection{Observation Weights and Their Gradients}
\label{app:relative_precision}

The bounded score transformation in Eq.~\eqref{eq:trust_decomposition}
uses $\ell_{\max}=\log2$. The normalization in
Eq.~\eqref{eq:relative_weight} finds $a_t$ by bisection, keeping visible
weights in $[0.5,2]$ with unit mean. Its active set is
$\mathcal U_t=\{p\in\mathcal V_t:0.5<w_{t,p}<2\}$.
Within an unchanged active set, differentiating the unit-mean constraint
gives the vector--Jacobian product
\begin{equation}
 \frac{\partial L}{\partial\ell_{t,p}}=
 \left\{\begin{array}{@{}l@{\;}l@{}}
 w_{t,p}\!\left(g_p-
 \dfrac{\sum_{q\in\mathcal U_t}w_{t,q}g_q}
       {\sum_{q\in\mathcal U_t}w_{t,q}}\right),&p\in\mathcal U_t,\\
 0,&\text{otherwise},
 \end{array}\right.
 \label{eq:joint_weight_vjp}
\end{equation}
where $g_p=\partial L/\partial w_{t,p}$. We use this implicit derivative
for backpropagation. Saturated and invisible observations have zero
derivatives. The regularizers assume at least one visible observation per
dynamic calibration frame.

\begin{figure*}[t]
  \centering
  \includegraphics[width=\textwidth]{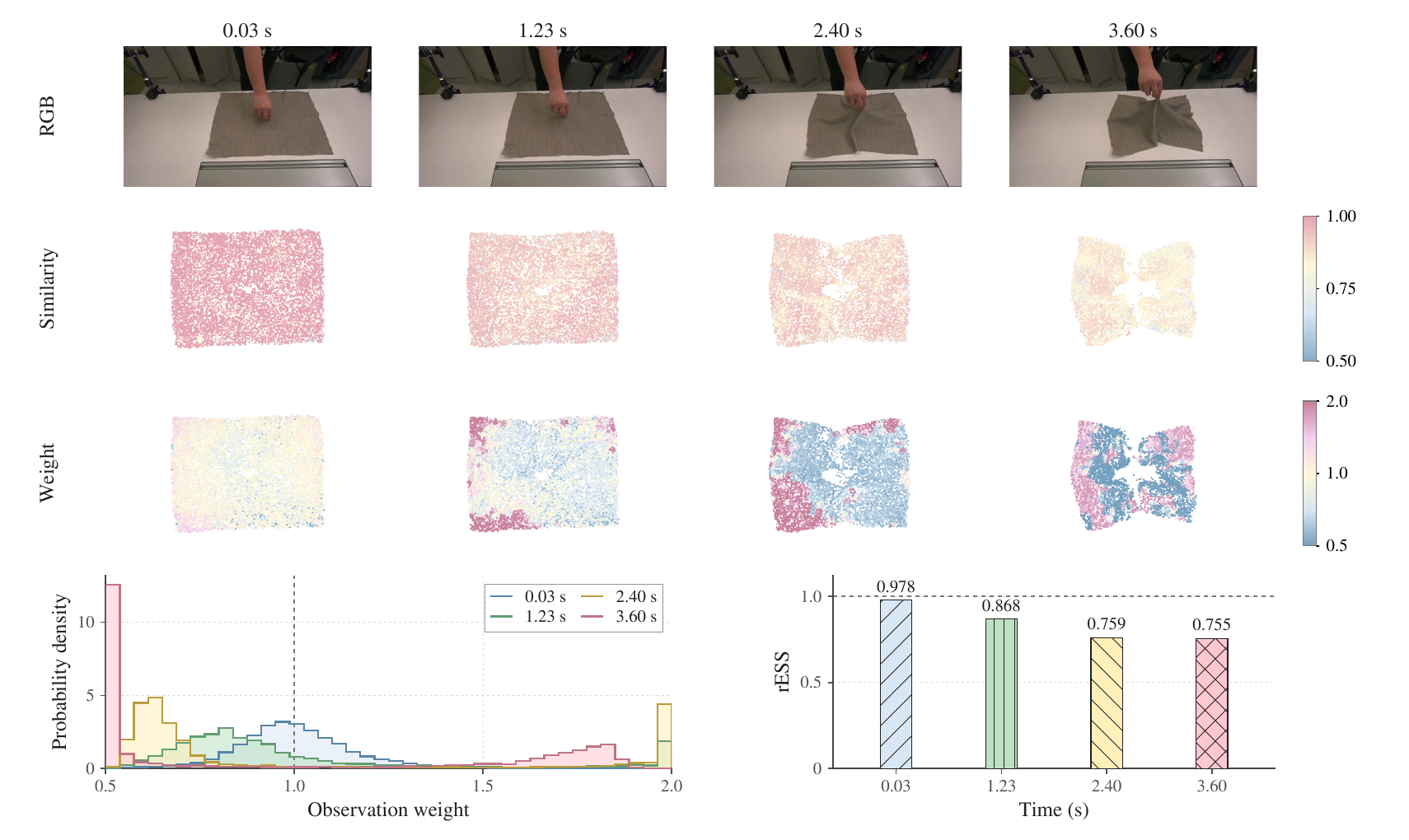}
  \caption{\textbf{Temporal evolution of single-stage observation trust on
  Lift Cloth 3.} Rows show RGB observations, frozen VideoMAE similarity, and
  learned pointwise weights at four calibration times. As deformation evolves,
  the weights become spatially differentiated beyond the raw similarity cue.
  Bottom: empirical probability densities of visible weights and relative
  effective sample size (rESS). The dashed line marks neutral influence
  ($w=1$). The distributions broaden from their initially concentrated profile,
  revealing down-weighted ($w<1$) and up-weighted ($w>1$) observations, while
  rESS stays above the prescribed 0.7 floor, indicating selectivity without
  weight collapse. Gray points are unavailable observations and are excluded
  from both statistics. No retraining, weight renormalization, or temporal
  interpolation is applied.}
  \label{fig:observation_trust_evolution}
\end{figure*}

Figure~\ref{fig:observation_trust_evolution} provides spatial and distributional
diagnostics of the bounded weights. A higher weight gives an observation
greater influence on physical fitting and should not be interpreted as a
contact probability. In the bottom-left panel, the horizontal axis is
observation weight and the vertical axis is its empirical probability density
across visible observations at each time. Lower density in a high-weight
interval means that fewer observations fall in that interval; each such
observation still receives greater influence in fitting. The curves do not
imply that density must decrease as weight increases. Their broadening over
time reflects increasing differentiation between down-weighted and up-weighted
observations.

\subsection{Initialization and Optimization}
\label{app:joint_implementation}

\paragraph{Initialization and optimizers.}
For each scene, we use the same CMA-initialized topology and physical
parameters as PhysTwin. The visual and contact score readouts have zero
weights and biases, yielding unit visible weights initially. All other adapter
layers use their standard random initialization, with random seed 42.
Following Algorithm~\ref{alg:v2_videomae}, we train for 200 epochs:
two physics-only warm-up epochs followed by 198 joint physics--adapter epochs.
Physical parameters use Adam with learning rate $10^{-3}$ and
$\beta=(0.9,0.99)$; the adapter uses AdamW with learning rate $10^{-4}$,
weight decay $10^{-3}$, and $\beta=(0.9,0.999)$.
Adapter gradients are clipped to a maximum norm of 5.
Each epoch starts from $\vect X_0$ and processes calibration frames in
temporal order.

\paragraph{Gradient computation.}
During joint training, at update $k$, weights are predicted with $\psi_k$ and the frame is
simulated with $\theta_k$. Write $\ell_t(\hat{\vect X},w)$ for the
reconstruction loss evaluated at predicted positions and observation
weights. The loss and regularizers are defined in
Appendix~\ref{app:observation_loss}. With $\operatorname{sg}$ denoting
stop-gradient, the two partial gradients are
\begin{equation}
 \begin{aligned}
 &g_\theta^k=\left.\nabla_\theta
     \ell_t(\hat{\vect X}_t(\theta),\operatorname{sg}(w_{\psi_k}))
     \right|_{\theta_k},\\
 &g_\psi^k=\left.\nabla_\psi\left[
     \ell_t(\operatorname{sg}(\hat{\vect X}_t^k),w_\psi)
     +s_tR_t(w_\psi)\right]\right|_{\psi_k}.
 \end{aligned}
 \label{eq:joint_partials}
\end{equation}
Both partial gradients are evaluated from the same simulation before
either parameter update. The simulated state is passed to the next frame
with gradients stopped, so physical differentiation is restricted to the
current frame.

\paragraph{Temporal adapter unrolling.}
At frame $t$, $\psi_k$ is rerun on the full prefix $0{:}t$ with zero GRU
state; no hidden state or graph is carried or detached across updates.
Both $w_{t-1}$ and $w_t$ are recomputed. Thus, the temporal penalty directly
couples only adjacent weights, whereas each adapter update back-propagates
through the complete $0{:}t$ GRU prefix. The computation graph is discarded
after the update and is not retained across optimizer steps.

\section{Physical Simulation and Joint Fitting}
\label{app:stage2}

This section expands the simulator in Sec.~\ref{sec:preliminaries} and
its role in the single-stage objective in Eq.~\eqref{eq:joint_objective}.

\subsection{Spring--Mass Transition}

We use the node and spring notation of Eq.~\eqref{eq:spring_force} and
omit the integration-substep index below.
The shared CMA initialization fixes the vertex set and masses, spring set
$\mathcal E_{\rm spr}$, rest lengths, controller links, collision distance, and
initial global parameters. Joint fitting represents
each spring stiffness by an unconstrained log parameter $\eta_s$ and evaluates
\begin{equation}
 k_s=\operatorname{clip}(\exp\eta_s,0,10^5).
 \label{eq:stiffness_parameterization}
\end{equation}
The spring force follows Eq.~\eqref{eq:spring_force}, with dashpot
coefficient $c_d$ fixed to its initialization. Controller endpoints
follow measured, linearly interpolated targets. The total force
$\vect f_i=\sum_{j\in\mathcal N(i)}\vect f_{ij}^{\rm spr}
+\vect f_i^{\rm ctrl}$ sums object-spring and controller-link forces.

For a substep of length $\Delta t=5\times10^{-5}$ s, force integration before
collision is
\begin{equation}
 \vect v_i^-=\exp(-c_a\Delta t)
 \left[\vect v_i+\Delta t
 \left(\frac{\vect f_i}{M_i}+\vect g\right)\right],
 \label{eq:velocity_update_app}
\end{equation}
where the air-drag coefficient $c_a$ is also fixed at initialization. This is
followed by object-contact impulses when enabled, then ground contact and a
position update. For a node crossing the ground with inward normal velocity,
the incident velocity is decomposed into normal and tangential components:
\begin{equation}
 \begin{aligned}
 &\vect v_n^+=-e_g\vect v_n^-,\quad
 \vect v_\tau^+=\beta_\tau\vect v_\tau^-,\\
 &\beta_\tau=\max\!\left(0,1-\mu_g\!(1+e_g)\!
 \frac{\|\vect v_n^-\|_2}
 {\max\!(\|\vect v_\tau^-\|_2,10^{-6})}\right).
 \end{aligned}
 \label{eq:ground_collision_app}
\end{equation}
The position update splits the substep at the ground time of impact,
using the incident velocity before impact and the resolved velocity after it.
Object--object response uses the analogous restitution and Coulomb tangential
impulse with the two masses. Effective restitution and friction are clipped
to $[0,1]$ and $[0,2]$, respectively, during simulation. Joint training optimizes
$\{\eta_s\}_{s=1}^{|\mathcal E_{\rm spr}|}$ together with
$(e_g,\mu_g,e_o,\mu_o)$; damping, masses, collision distance, topology, and
controller motion remain fixed. There are 667 substeps per 30-FPS video frame.

\subsection{Weighted Observation Loss}
\label{app:observation_loss}
\label{app:checkpoint_selection}

\paragraph{Reconstruction terms.}
We use the reconstruction terms and valid-point normalization in
Eq.~\eqref{eq:joint_reconstruction}, with a unit SmoothL1 transition threshold.

\paragraph{Anti-collapse terms and scale.}
With $n_t=|\mathcal V_t|$ and
$\mathcal B_t=\mathcal V_t\cap\mathcal V_{t-1}$, the regularizers are
\begin{equation}
 \begin{aligned}
 &R_{{\rm KL},t}=\frac1{n_t}\sum_{p\in\mathcal V_t}
                   (w_{t,p}\log w_{t,p}-w_{t,p}+1),\\
 &R_{{\rm temp},t}=\frac{\sum_{p\in\mathcal B_t}
                          (w_{t,p}-w_{t-1,p})^2}
                         {\max(1,|\mathcal B_t|)},\\
 &r_t=\frac{(\sum_{p\in\mathcal V_t}w_{t,p})^2}
                 {n_t\sum_{p\in\mathcal V_t}w_{t,p}^2}.
 \end{aligned}
 \label{eq:joint_weight_regularizers}
\end{equation}
The visible unit mean makes $R_{{\rm KL},t}$ the KL divergence of normalized
influence from the uniform distribution. We use
$R_{{\rm ESS},t}=[\max(0,0.7-r_t)]^2$ and
$(\lambda_{\rm KL},\lambda_{\rm temp},\lambda_{\rm ESS})=(1,0.1,10)$.
Let $\mathcal L_{{\rm rec},t}^{\mathbf1}$ use unit weights on valid points.
The detached scale is
\begin{equation}
 s_t=\operatorname{sg}\!\left[
          \max(\mathcal L_{{\rm rec},t}^{\mathbf1},10^{-8})\right],
 \label{eq:joint_residual_scale}
\end{equation}
where $\operatorname{sg}$ stops gradients.

\paragraph{Calibration-loss checkpoint selection.}
For epoch $e$, let $\mathcal L_{{\rm rec},t}^{(e)}$ denote the weighted
reconstruction loss at frame $t$, evaluated before the parameter update.
We select the saved checkpoint minimizing its mean over the calibration
prefix:
\begin{equation}
 \begin{aligned}
 &\overline{\mathcal L}_{\rm cal}^{(e)}=
 \frac{1}{T_c-1}\sum_{t=1}^{T_c-1}\mathcal L_{{\rm rec},t}^{(e)},\\
 &e^*=\arg\min_{e\in\mathcal E_{\rm save}}
              \overline{\mathcal L}_{\rm cal}^{(e)}.
 \end{aligned}
 \label{eq:calibration_checkpoint_selection}
\end{equation}
Here, $\mathcal E_{\rm save}$ denotes the saved training epochs.
Selection uses reconstruction loss alone, excluding regularization terms.

\subsection{Gaussian Motion Transfer}
\label{app:gaussian_skinning}

We transfer the simulated motion to the canonical Gaussian representation
$\mathcal A_0$ using the geometric skinning procedure inherited from
PhysTwin~\cite{jiang2025phystwin}. The simulation nodes and Gaussians share
the same world coordinate system. Here, $i$ indexes spring--mass nodes and
$a$ indexes Gaussians. Gaussian $a$ has center $\vect\mu_{t,a}$, unit
orientation quaternion $\vect q^{\rm G}_{t,a}$, scale $\vect s_a$, opacity,
and appearance coefficients. The transfer updates its center and orientation
from consecutive simulated states $\vect X_t$ and $\vect X_{t+1}$.

\paragraph{Local motion of simulation nodes.}
For each node $i$, we select its 16 nearest other simulation nodes at
frame zero, forming a fixed neighborhood $\mathcal N_i^{\rm rot}$.
Let $\vect d_{t,ij}=\vect x_{t,j}-\vect x_{t,i}$ be a neighbor's relative
position. We estimate the incremental rotation $\vect R_{t,i}$ by
least-squares alignment of these offsets between consecutive frames.
The cross-covariance is
\begin{equation}
 \vect H_{t,i}=\sum_{j\in\mathcal N_i^{\rm rot}}
 \vect d_{t+1,ij}\vect d_{t,ij}^{\top}.
 \label{eq:node_rotation_covariance_app}
\end{equation}
Writing its singular value decomposition as
$\vect H_{t,i}=\vect U\vect\Lambda\vect V^\top$, the rotation is
\begin{equation}
 \vect R_{t,i}=\vect U\,
 \operatorname{diag}\!\left(1,1,\det(\vect U\vect V^\top)\right)
 \vect V^\top.
 \label{eq:node_rotation_svd_app}
\end{equation}
The determinant correction ensures a proper rotation. Together with the
node displacement $\vect x_{t+1,i}-\vect x_{t,i}$, this defines a local
rigid motion around node $i$.

\paragraph{Gaussian-to-node association.}
For Gaussian center $\vect\mu_{t,a}$, let $\mathcal K_{t,a}$ contain its
16 nearest simulation nodes at frame $t$. This association is recomputed
from the current Gaussian and node positions at every frame. The influence
of each neighboring node is given by normalized inverse-distance weights:
\begin{equation}
 b_{t,ai}=\frac{(\|\vect\mu_{t,a}-\vect x_{t,i}\|_2+10^{-6})^{-1}}
 {\sum_{j\in\mathcal K_{t,a}}
 (\|\vect\mu_{t,a}-\vect x_{t,j}\|_2+10^{-6})^{-1}}.
 \label{eq:skinning_weights_app}
\end{equation}
Thus, $b_{t,ai}\geq0$ and $\sum_i b_{t,ai}=1$, with greater influence
assigned to closer nodes. The node neighborhoods $\mathcal N_i^{\rm rot}$
remain fixed for rotation estimation, while the Gaussian neighborhoods
$\mathcal K_{t,a}$ follow the evolving geometry. These geometric skinning
weights are independent of the learned observation weights $\vect W_\psi$.

\paragraph{Center update.}
Each neighboring node proposes a new Gaussian center by rotating its
current offset from that node and translating it to the node's next
position. Blending these proposals gives
\begin{equation}
 \vect\mu_{t+1,a}=\sum_{i\in\mathcal K_{t,a}}b_{t,ai}
 [\vect R_{t,i}(\vect\mu_{t,a}-\vect x_{t,i})+\vect x_{t+1,i}].
 \label{eq:gaussian_center_update_app}
\end{equation}
Without rotation, the center moves by the weighted average of neighboring
node displacements.

\paragraph{Orientation and covariance update.}
Let $\delta\vect q_{t,i}$ be the unit quaternion corresponding to
$\vect R_{t,i}$. We blend the incremental rotations using the same skinning
weights and compose the result with the Gaussian's current orientation:
\begin{equation}
 \begin{aligned}
 &\delta\vect q^{\rm G}_{t,a}=\operatorname{norm}_2\!\left(
 \sum_{i\in\mathcal K_{t,a}}b_{t,ai}\,\delta\vect q_{t,i}\right),\\
 &\vect q^{\rm G}_{t+1,a}
 =\delta\vect q^{\rm G}_{t,a}\otimes\vect q^{\rm G}_{t,a}.
 \end{aligned}
 \label{eq:gaussian_orientation_update_app}
\end{equation}
Here, $\operatorname{norm}_2$ normalizes a quaternion to unit length and
$\otimes$ denotes quaternion multiplication. Left multiplication applies
the incremental world-space rotation to the existing orientation. With
$\vect Q_{t+1,a}$ the rotation matrix of $\vect q^{\rm G}_{t+1,a}$, the
updated Gaussian covariance is
\begin{equation}
 \vect\Sigma_{t+1,a}=\vect Q_{t+1,a}
 \operatorname{diag}(\vect s_a^2)\vect Q_{t+1,a}^{\top},
 \label{eq:gaussian_covariance_update_app}
\end{equation}
where the square is componentwise. Scales, opacities, and appearance
coefficients remain fixed, so object deformation is represented by the
evolving Gaussian centers and orientations.

Starting from $\mathcal A_0$, we sequentially update and render the
Gaussians by calibrated projection and alpha compositing
(Eq.~\eqref{eq:physics_rendering}). This geometric transfer introduces
no learned parameters.

\section{Evaluation Protocol}
\label{app:evaluation}

For Spring-Gaus~\cite{zhong2024springgaus},
GS-Dynamics~\cite{zhang2024gsdynamics}, PhysTwin~\cite{jiang2025phystwin},
and MatPhys~\cite{yang2026matphys}, we follow the official released
implementations, adapting data interfaces and calibration schedules
to our benchmark. MatPhys uses the released epoch-200 checkpoint and
uniform-material fallback and is reported as a pretrained-model reference
because its original training protocol differs from our temporal-prefix
setting. For DeformMaster~\cite{li2026deformmaster}, whose evaluated release
does not include full training code or configurations, we reproduce the training pipeline from
the paper using the released components. Specifically, we implement
differentiable MPM training, controller-gain optimization, and RGB refinement,
with calibration-only adjustments to residual learning rates and MPM time
steps for stability while preserving frame duration and the temporal gradient
horizon. Data splits and metrics follow Sec.~\ref{sec:experiments}; geometry
and rendering cover all 22 scenes and 735 future frames, while tracking uses
732 frames after excluding three without valid annotations (Lift Dinosaur,
83--85).

\clearpage
\onecolumn
\section{Complete Per-Scene Results and Ablations}
\label{app:complete_results}
\label{app:heuristic_weighting}

Tables~\ref{tab:per_scene_reconstruction} and~\ref{tab:per_scene_physics}
give per-scene baseline results; Tables~\ref{tab:per_scene_ablation_reconstruction}
and~\ref{tab:per_scene_ablation_prediction} give ablations, including
CoTracker confidence and residual weighting. Together they
report all nine metrics on all 22 scenes for reconstruction/resimulation
and future prediction. Overall, \method outperforms the evaluated state-of-the-art baselines
in future prediction.

\textbf{Parameter-free weighting.}
CoTracker confidence uses the product of CoTracker3's~\cite{karaev2024cotracker3}
continuous visibility and confidence probabilities, re-extracted on calibration
frames without changing the original 3D observations or masks.
Residual weighting uses detached Cauchy scores $[1+(r/s)^2]^{-1}$,
where $r$ is the identity-aligned 3D residual and $s$ its visible-point
median, floored at 1 mm. Both omit the adapter, contact graph, and GRU
and retain bounded $[0.5,2]$, unit-mean weights. They match the separate
single-stage weight-regularization reference (Sec.~\ref{sec:ablation}),
including initialization, simulator, optimizer, 200 epochs, two uniform-weight
warm-up epochs, and epoch selection.
Bold and underlined values rank the six learned-weight rows6.

\begingroup
\footnotesize
\setlength{\tabcolsep}{2.0pt}
\setlength{\LTleft}{\fill}\setlength{\LTright}{\fill}
\begin{longtable}{llrrrrrrrrr}
\caption{\textbf{Per-scene baseline results: reconstruction and resimulation.} Each scene includes five baselines and the full ViTrust model (shaded). Distances are in mm; PSNR in dB; SSIM, LPIPS, and IoU are scaled by $100$. Each method follows its native protocol. Baseline settings are in Appendix~\ref{app:evaluation}. Best and second-best numerical values among the six displayed rows within each scene are bold and underlined, using unrounded values.}\label{tab:per_scene_reconstruction}\\
\toprule
\multicolumn{2}{l}{\cellcolor{reconstructiongreen}Reconstruction \& Resimulation} & \multicolumn{5}{c}{Physical (mm)} & \multicolumn{4}{c}{Rendering (PSNR: dB; others: $\times100$)}\\
\cmidrule(lr){3-7}\cmidrule(lr){8-11}
Scene & Method / Variant & CD $\downarrow$ & Track $\downarrow$ & RMSE $\downarrow$ & Median $\downarrow$ & P95 $\downarrow$ & PSNR $\uparrow$ & SSIM $\uparrow$ & LPIPS $\downarrow$ & IoU $\uparrow$\\
\midrule
\endfirsthead
\toprule
\multicolumn{2}{l}{\cellcolor{reconstructiongreen}Reconstruction \& Resimulation} & \multicolumn{5}{c}{Physical (mm)} & \multicolumn{4}{c}{Rendering (PSNR: dB; others: $\times100$)}\\
\cmidrule(lr){3-7}\cmidrule(lr){8-11}
Scene & Method / Variant & CD $\downarrow$ & Track $\downarrow$ & RMSE $\downarrow$ & Median $\downarrow$ & P95 $\downarrow$ & PSNR $\uparrow$ & SSIM $\uparrow$ & LPIPS $\downarrow$ & IoU $\uparrow$\\
\midrule
\endhead
\bottomrule
\endfoot
Cloth Lift 1 (C) & Spring-Gaus (ECCV 2024) & 10.64 & 10.54 & 12.37 & 8.85 & 20.08 & 27.59 & 95.15 & 4.85 & \underline{91.80}\\*
 & GS-Dynamics (CoRL 2024) & 12.48 & 25.60 & 26.67 & 25.61 & 36.11 & 24.28 & 94.44 & 7.47 & 80.92\\*
 & PhysTwin (ICCV 2025) & \underline{3.04} & \best{3.33} & \best{4.12} & \underline{2.42} & \underline{7.29} & 27.57 & \underline{95.19} & \underline{3.58} & 91.65\\*
 & MatPhys (arXiv 2026) & 3.92 & 8.39 & 8.65 & 8.50 & 10.84 & 27.41 & 95.13 & 4.08 & 91.48\\*
 & DeformMaster (NeurIPS 2026) & 3.33 & 4.26 & 4.59 & 4.22 & \best{6.71} & \best{32.44} & \best{97.82} & \best{2.30} & \best{95.35}\\*
\rowcolor{tableoneblue} & ViTrust (ours) & \best{3.02} & \underline{3.33} & \underline{4.15} & \best{2.42} & 7.33 & \underline{27.66} & 95.17 & 3.65 & 91.74\\
\addlinespace[1.5pt]
Cloth Lift 3 (C) & Spring-Gaus (ECCV 2024) & 6.54 & 13.89 & 16.44 & 10.87 & 28.19 & \underline{26.81} & 87.93 & 7.87 & 92.64\\*
 & GS-Dynamics (CoRL 2024) & 19.47 & 33.95 & 35.00 & 30.99 & 47.87 & 23.61 & 87.24 & 12.79 & 85.00\\
\end{longtable}

\clearpage
\footnotesize
\setlength{\tabcolsep}{2.0pt}
\setlength{\LTleft}{\fill}\setlength{\LTright}{\fill}
\begin{longtable}{llrrrrrrrrr}
\caption{\textbf{Per-scene baseline results: future prediction.} Each scene includes five baselines and the full ViTrust model (shaded). Distances are in mm; PSNR in dB; SSIM, LPIPS, and IoU are scaled by $100$. Baseline settings are in Appendix~\ref{app:evaluation}. Best and second-best numerical values among the six displayed rows within each scene are bold and underlined, using unrounded values. Lift Dinosaur tracking metrics use the same 23 annotated future frames as ViTrust; CD and rendering use all 26 frames.}\label{tab:per_scene_physics}\\
\toprule
\multicolumn{2}{l}{\cellcolor{predictionpink}Future Prediction} & \multicolumn{5}{c}{Physical (mm)} & \multicolumn{4}{c}{Rendering (PSNR: dB; others: $\times100$)}\\
\cmidrule(lr){3-7}\cmidrule(lr){8-11}
Scene & Method / Variant & CD $\downarrow$ & Track $\downarrow$ & RMSE $\downarrow$ & Median $\downarrow$ & P95 $\downarrow$ & PSNR $\uparrow$ & SSIM $\uparrow$ & LPIPS $\downarrow$ & IoU $\uparrow$\\
\midrule
\endfirsthead
\toprule
\multicolumn{2}{l}{\cellcolor{predictionpink}Future Prediction} & \multicolumn{5}{c}{Physical (mm)} & \multicolumn{4}{c}{Rendering (PSNR: dB; others: $\times100$)}\\
\cmidrule(lr){3-7}\cmidrule(lr){8-11}
Scene & Method / Variant & CD $\downarrow$ & Track $\downarrow$ & RMSE $\downarrow$ & Median $\downarrow$ & P95 $\downarrow$ & PSNR $\uparrow$ & SSIM $\uparrow$ & LPIPS $\downarrow$ & IoU $\uparrow$\\
\midrule
\endhead
\bottomrule
\endfoot
Cloth Lift 1 (C) & Spring-Gaus (ECCV 2024) & 18.20 & 18.00 & 21.70 & \underline{13.68} & 38.91 & 25.49 & 94.18 & 7.02 & 85.32\\*
 & GS-Dynamics (CoRL 2024) & 15.93 & 38.98 & 53.97 & 31.69 & 98.08 & 20.66 & 92.52 & 12.76 & 62.36\\*
 & PhysTwin (ICCV 2025) & 10.81 & 26.95 & 28.75 & 23.73 & 42.24 & 23.63 & 93.76 & 6.77 & 82.75\\*
 & MatPhys (arXiv 2026) & \underline{7.59} & 17.26 & 17.98 & 16.92 & 24.55 & 24.95 & 94.03 & 6.24 & 86.73\\*
 & DeformMaster (NeurIPS 2026) & 11.45 & \underline{15.62} & \underline{16.69} & 15.17 & \best{24.10} & \underline{26.13} & \best{96.04} & \underline{5.36} & \underline{87.51}\\*
\rowcolor{tableoneblue} & ViTrust (ours) & \best{4.98} & \best{9.76} & \best{12.84} & \best{6.52} & \underline{24.12} & \best{27.37} & \underline{94.77} & \best{4.01} & \best{93.51}\\
\addlinespace[1.5pt]
Cloth Lift 3 (C) & Spring-Gaus (ECCV 2024) & 11.48 & 52.64 & 67.55 & 36.68 & 118.23 & 24.81 & 86.73 & 10.25 & 89.06\\*
 & GS-Dynamics (CoRL 2024) & 24.47 & 115.46 & 131.84 & 103.90 & 212.64 & 18.02 & 83.25 & 23.00 & 51.20\\
\end{longtable}

\clearpage
\footnotesize
\setlength{\tabcolsep}{2.0pt}
\setlength{\LTleft}{\fill}\setlength{\LTright}{\fill}
\renewcommand{\arraystretch}{0.96}
\begin{longtable}{llrrrrrrrrr}
\caption{\textbf{Per-scene ablation results: reconstruction and resimulation.} Each scene includes seven ablation variants and the full ViTrust model (shaded). Distances are in mm; PSNR in dB; SSIM, LPIPS, and IoU are scaled by $100$. Definitions and protocol differences are given in Appendix~\ref{app:complete_results}. Best and second-best numerical values among the six learned-weight rows within each scene are bold and underlined, using unrounded values.}\label{tab:per_scene_ablation_reconstruction}\\
\toprule
\multicolumn{2}{l}{\cellcolor{reconstructiongreen}Reconstruction \& Resimulation} & \multicolumn{5}{c}{Physical (mm)} & \multicolumn{4}{c}{Rendering (PSNR: dB; others: $\times100$)}\\
\cmidrule(lr){3-7}\cmidrule(lr){8-11}
Scene & Method / Variant & CD $\downarrow$ & Track $\downarrow$ & RMSE $\downarrow$ & Median $\downarrow$ & P95 $\downarrow$ & PSNR $\uparrow$ & SSIM $\uparrow$ & LPIPS $\downarrow$ & IoU $\uparrow$\\
\midrule
\endfirsthead
\toprule
\multicolumn{2}{l}{\cellcolor{reconstructiongreen}Reconstruction \& Resimulation} & \multicolumn{5}{c}{Physical (mm)} & \multicolumn{4}{c}{Rendering (PSNR: dB; others: $\times100$)}\\
\cmidrule(lr){3-7}\cmidrule(lr){8-11}
Scene & Method / Variant & CD $\downarrow$ & Track $\downarrow$ & RMSE $\downarrow$ & Median $\downarrow$ & P95 $\downarrow$ & PSNR $\uparrow$ & SSIM $\uparrow$ & LPIPS $\downarrow$ & IoU $\uparrow$\\
\midrule
\endhead
\bottomrule
\endfoot
Cloth Lift 1 (C) & Visual trust only & 4.39 & 11.66 & 11.81 & 11.44 & 14.48 & \best{31.18} & \best{95.96} & \best{2.89} & \best{95.58}\\*
 & Contact-graph trust only & \underline{3.02} & \best{3.32} & \best{4.14} & \underline{2.43} & \best{7.29} & 27.68 & 95.17 & 3.68 & 91.75\\*
 & Frame-wise (no GRU) & 3.07 & 3.34 & 4.16 & 2.45 & 7.33 & 27.69 & 95.17 & 3.67 & 91.75\\*
 & w/o Weight regularization & 3.73 & 5.29 & 5.86 & 4.76 & 9.19 & 26.53 & 94.88 & 4.12 & 89.61\\*
 & CoTracker confidence & 3.21 & 3.63 & 4.43 & 2.90 & 7.79 & 27.29 & 95.16 & 3.68 & 91.01\\*
 & Residual weighting & 3.63 & 4.50 & 5.17 & 3.88 & 8.59 & 26.87 & 94.99 & 3.92 & 90.34\\*
 & w/ DINO & 3.05 & 3.34 & \underline{4.14} & 2.45 & \underline{7.30} & \underline{27.73} & \underline{95.18} & 3.70 & \underline{91.78}\\*
\rowcolor{tableoneblue} & ViTrust (ours) & \best{3.02} & \underline{3.33} & 4.15 & \best{2.42} & 7.33 & 27.66 & 95.17 & \underline{3.65} & 91.74\\
\end{longtable}

\clearpage
\footnotesize
\setlength{\tabcolsep}{2.0pt}
\setlength{\LTleft}{\fill}\setlength{\LTright}{\fill}
\begin{longtable}{llrrrrrrrrr}
\caption{\textbf{Per-scene ablation results: future prediction.} Each scene includes seven ablation variants and the full ViTrust model (shaded). Distances are in mm; PSNR in dB; SSIM, LPIPS, and IoU are scaled by $100$. Definitions and protocol differences are given in Appendix~\ref{app:complete_results}. Best and second-best numerical values among the six learned-weight rows within each scene are bold and underlined, using unrounded values. Lift Dinosaur tracking metrics use the same 23 annotated future frames as ViTrust; CD and rendering use all 26 frames.}\label{tab:per_scene_ablation_prediction}\\
\toprule
\multicolumn{2}{l}{\cellcolor{predictionpink}Future Prediction} & \multicolumn{5}{c}{Physical (mm)} & \multicolumn{4}{c}{Rendering (PSNR: dB; others: $\times100$)}\\
\cmidrule(lr){3-7}\cmidrule(lr){8-11}
Scene & Method / Variant & CD $\downarrow$ & Track $\downarrow$ & RMSE $\downarrow$ & Median $\downarrow$ & P95 $\downarrow$ & PSNR $\uparrow$ & SSIM $\uparrow$ & LPIPS $\downarrow$ & IoU $\uparrow$\\
\midrule
\endfirsthead
\toprule
\multicolumn{2}{l}{\cellcolor{predictionpink}Future Prediction} & \multicolumn{5}{c}{Physical (mm)} & \multicolumn{4}{c}{Rendering (PSNR: dB; others: $\times100$)}\\
\cmidrule(lr){3-7}\cmidrule(lr){8-11}
Scene & Method / Variant & CD $\downarrow$ & Track $\downarrow$ & RMSE $\downarrow$ & Median $\downarrow$ & P95 $\downarrow$ & PSNR $\uparrow$ & SSIM $\uparrow$ & LPIPS $\downarrow$ & IoU $\uparrow$\\
\midrule
\endhead
\bottomrule
\endfoot
Cloth Lift 1 (C) & Visual trust only & 9.38 & 24.66 & 25.23 & 23.27 & 33.32 & 25.20 & 94.16 & 6.09 & 87.04\\*
 & Contact-graph trust only & 5.09 & \underline{9.61} & \underline{12.70} & \best{6.05} & \underline{23.94} & 27.12 & 94.72 & 4.12 & 93.15\\*
 & Frame-wise (no GRU) & 7.73 & 16.02 & 18.35 & 13.17 & 30.96 & 25.47 & 94.38 & 4.98 & 89.29\\*
 & w/o Weight regularization & 9.25 & 21.68 & 23.75 & 18.78 & 37.18 & 24.46 & 94.20 & 5.83 & 86.54\\*
 & CoTracker confidence & 8.81 & 19.42 & 21.66 & 17.91 & 35.23 & 24.85 & 94.29 & 5.36 & 87.63\\*
 & Residual weighting & 8.74 & 18.57 & 20.91 & 15.95 & 34.57 & 24.97 & 94.36 & 5.35 & 88.18\\*
 & w/ DINO & \best{4.93} & \best{9.54} & \best{12.56} & \underline{6.19} & \best{23.65} & \best{27.42} & \best{94.83} & \best{3.96} & \best{93.66}\\*
\rowcolor{tableoneblue} & ViTrust (ours) & \underline{4.98} & 9.76 & 12.84 & 6.52 & 24.12 & \underline{27.37} & \underline{94.77} & \underline{4.01} & \underline{93.51}\\
\end{longtable}

\endgroup

\end{document}